%% file: main.tex
\documentclass[10pt,twocolumn,letterpaper]{article}

\usepackage[pagenumbers]{cvpr} 

\input{preamble}
\definecolor{cvprblue}{rgb}{0.21,0.49,0.74}
\usepackage[pagebackref,breaklinks,colorlinks,allcolors=cvprblue]{hyperref}
\usepackage{amsmath,amsfonts}
\usepackage{algorithm}
\usepackage{algorithmic}
\usepackage{adjustbox}
\usepackage{booktabs}
\usepackage{multirow}
\usepackage{color}
\usepackage{float} 
\usepackage[table]{xcolor}
\def\paperID{3036} 
\def\confName{CVPR}
\def\confYear{2026}

\title{Reclaiming Lost Text Layers for Source-Free Cross-Domain Few-Shot Learning}

\author
  {Zhenyu Zhang$^{1}$,\quad Guangyao Chen$^{2}$\thanks{Corresponding author.},\quad Yixiong Zou$^{1}$\footnotemark[1],\quad Yuhua Li$^{1}$\footnotemark[1],\quad Ruixuan Li$^{1}$\\ 
  $^{1}$ School of Computer Science and Technology, Huazhong University of Science and Technology \\
  $^{2}$ National Key Laboratory for Multimedia Information Processing,  Peking University\\ 
  {\tt\small d202481641hust.edu.cn, gy.chen@pku.edu.cn, \{yixiongz, idcliyuhua, rxli\}@hust.edu.cn}
  }

\begin{document}
\maketitle
\input{./Tex/00_abs_yxz}
\input{./Tex/01_intro_yxz}

\input{./Tex/02_related_yxz}
\input{./Tex/03_preliminary_yxz}
\input{./Tex/04_ana_yxz}

\input{./Tex/05_method_yxz}

\input{./Tex/06_exp_yxz}

\input{./Tex/07_con}
\section{Acknowledgments}
This work is supported by the National Key Research and Development Program of China under grant 2024YFC3307900; the National Natural Science Foundation of China under grants 62436003, 62206102, 62376103, 62302184, and 62402015; the Major Science and Technology Project of Hubei Province under grant 2025BAB011 and 2024BAA008; the Hubei Science and Technology Talent Service Project under grant 2024DJC078; Ant Group through the CCF–Ant Research Fund; the Postdoctoral Fellowship Program of the China Postdoctoral Science Foundation under grant GZB20230024; and the China Postdoctoral Science Foundation under grant 2024M750100. Computations were performed on the HPC Platform of Huazhong University of Science and Technology.
{
    \small
    \bibliographystyle{ieeenat_fullname}
    \bibliography{main}
}

\newpage

\noindent{\Large \textbf{Appendix}}
\tableofcontents
\input{./Tex/08_sup}


\end{document}

%% file: Tex/00_abs_yxz.tex
\begin{abstract}

Source-Free Cross-Domain Few-Shot Learning (SF-CDFSL) focuses on fine-tuning with limited training data from target domains (e.g., medical or satellite images), where CLIP has recently shown promising results
due to its generalizability to downstream tasks. 
Current works indicate CLIP's text encoder is more suitable for cross-domain tasks, however, we find that \textbf{removing certain middle layers of the text encoder can effectively improve performance in SF-CDFSL}, which we call the Lost Layers. In this paper, we delve into this phenomenon for a deeper understanding. We discover that instead of being harmful for the SF-CDFSL task, the information in these layers is actually beneficial, but visual gaps prevent this useful information from being fully utilized, making these layers seem redundant.
Based on this understanding, unlike current works that simply remove these layers, we propose a method to 
teachs the model to \textbf{re-utilize} information in these lost layers
at both the layer and encoder levels, guiding the re-learning of the visual branch under domain shifts. Our approach effectively addresses the issue of underutilized information in the text encoder.
Extensive experiments across various settings, backbones (CLIP, SigLip, PE-Core), and tasks (4 CDFSL datasets and 10 Meta-dataset datasets) demonstrate the effectiveness of our method. Code is available at https://github.com/zhenyuZ-HUST/CVPR26-VtT.

\end{abstract}

%% file: Tex/01_intro_yxz.tex
\section{Introduction}
\begin{figure*}[!h]
\centering
\subfloat[CLIP]{
\centering
\label{fig:intro_clip}
\begin{minipage}[b]{0.49\linewidth}
\begin{adjustbox}{max width=1\linewidth}
\includegraphics[]{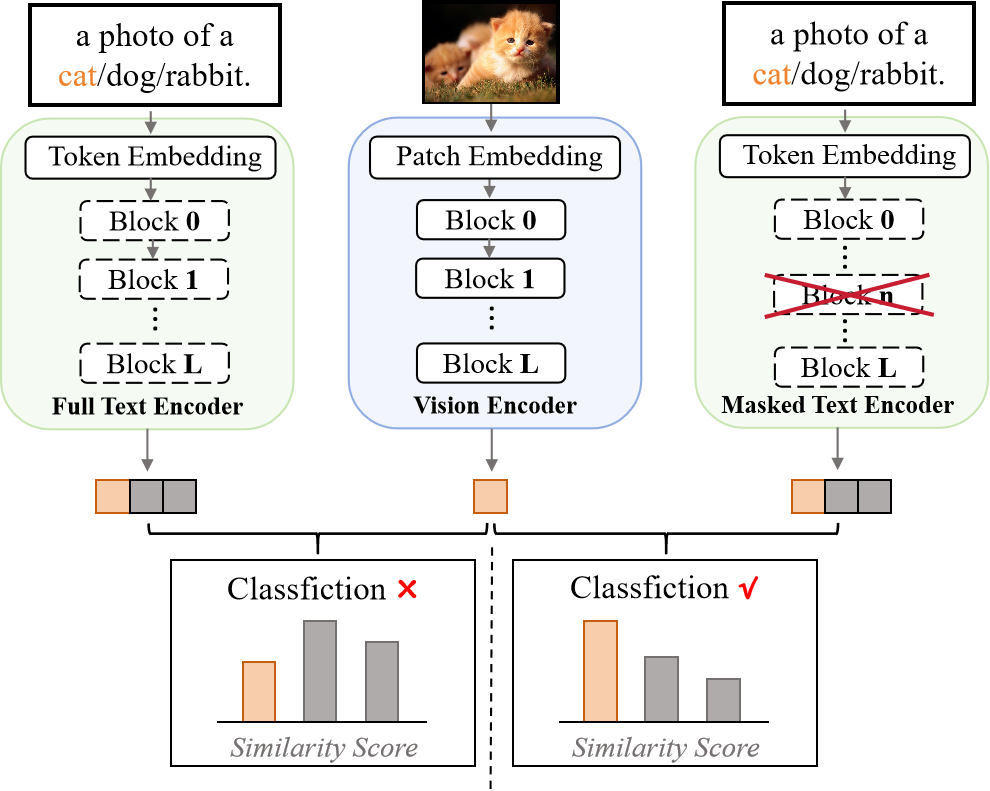}
\end{adjustbox}
\end{minipage}
} 
\subfloat[Base]{
\label{fig:intro_lost_base}
\begin{minipage}[b]{0.23\linewidth}
\includegraphics[width=1\linewidth]{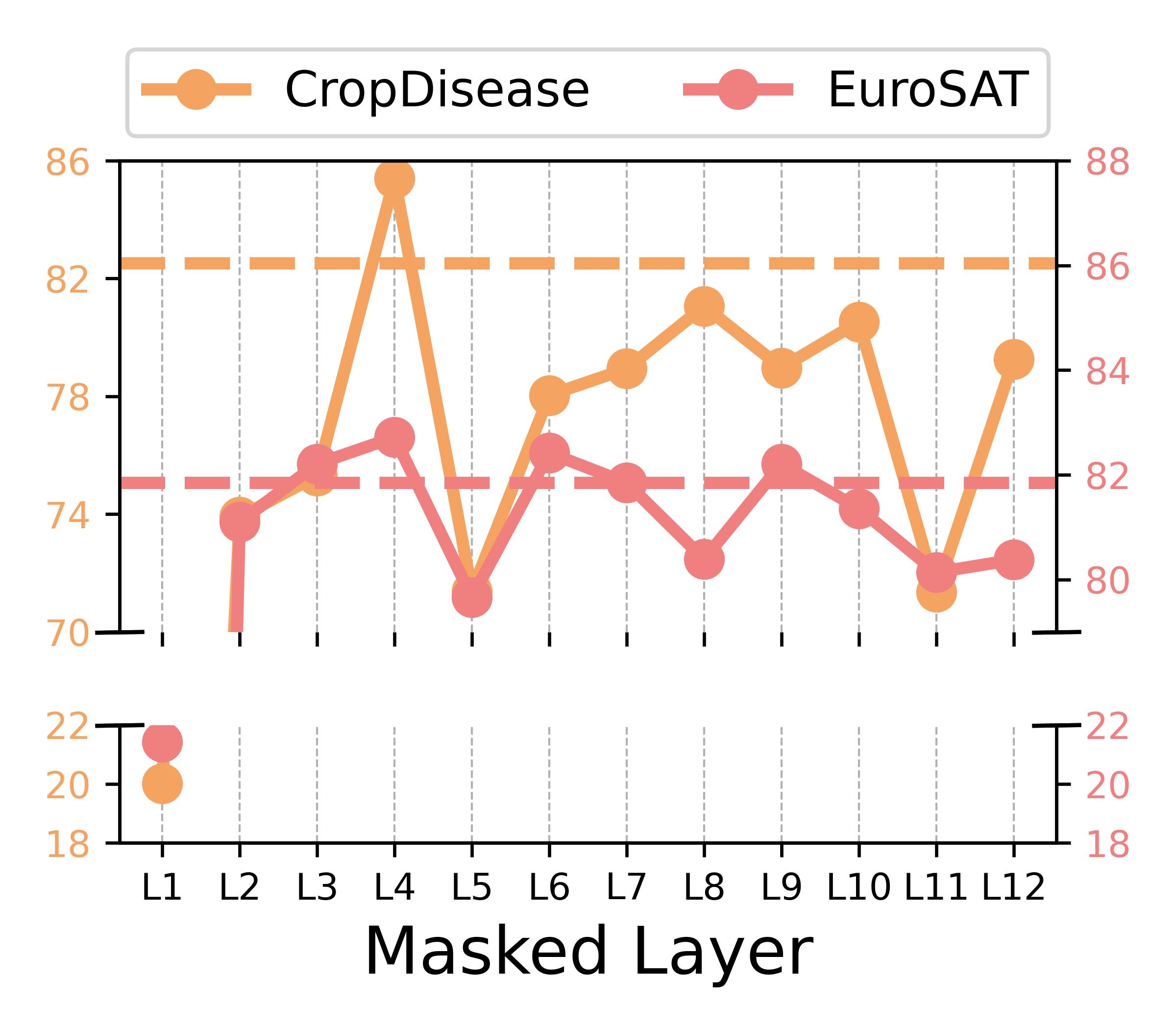}
\\
\includegraphics[width=1\linewidth]{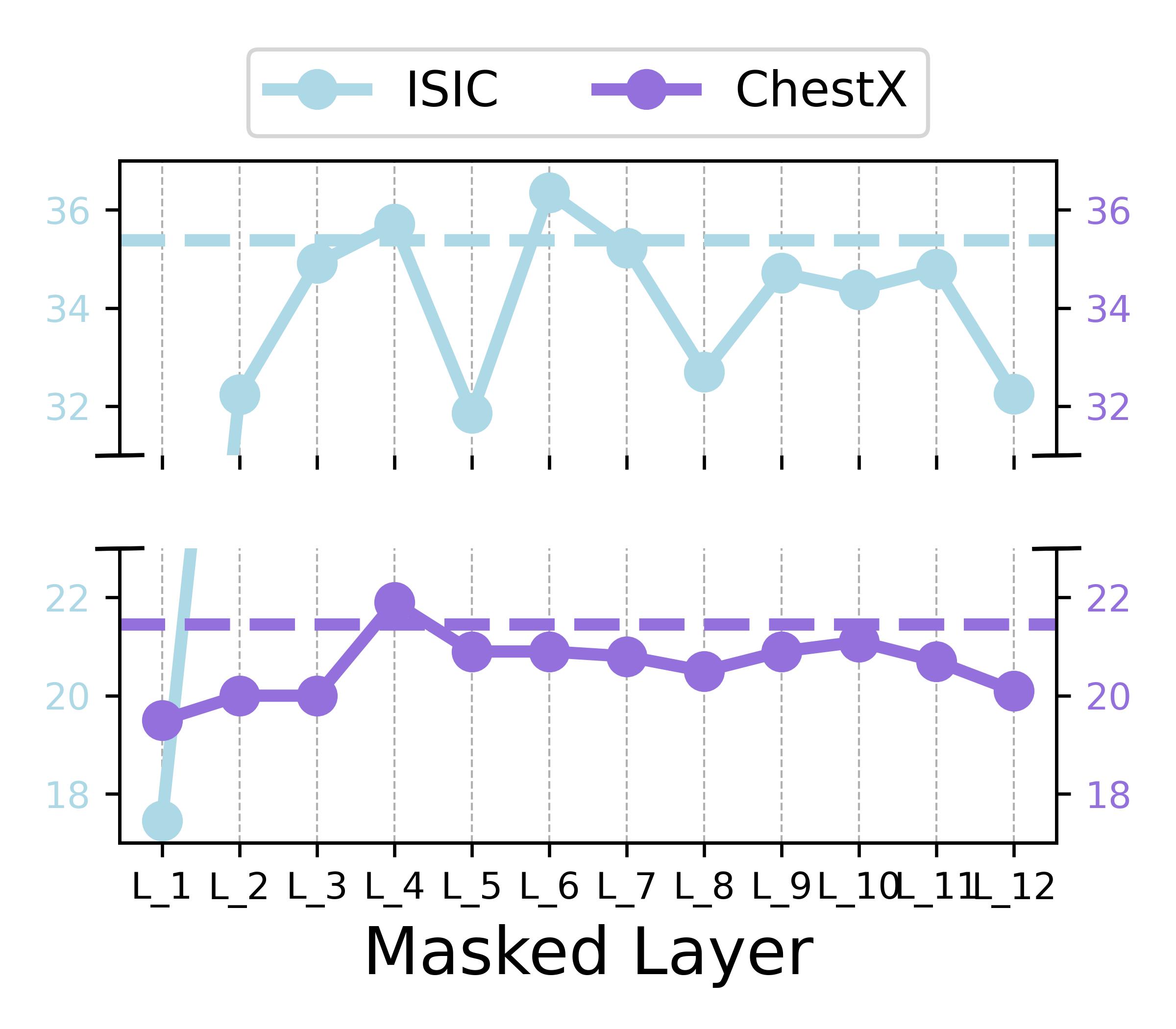}
\end{minipage}
}
\subfloat[Ours]{
\label{fig:intro_lost_ours}
\begin{minipage}[b]{0.23\linewidth}
\includegraphics[width=1\linewidth]{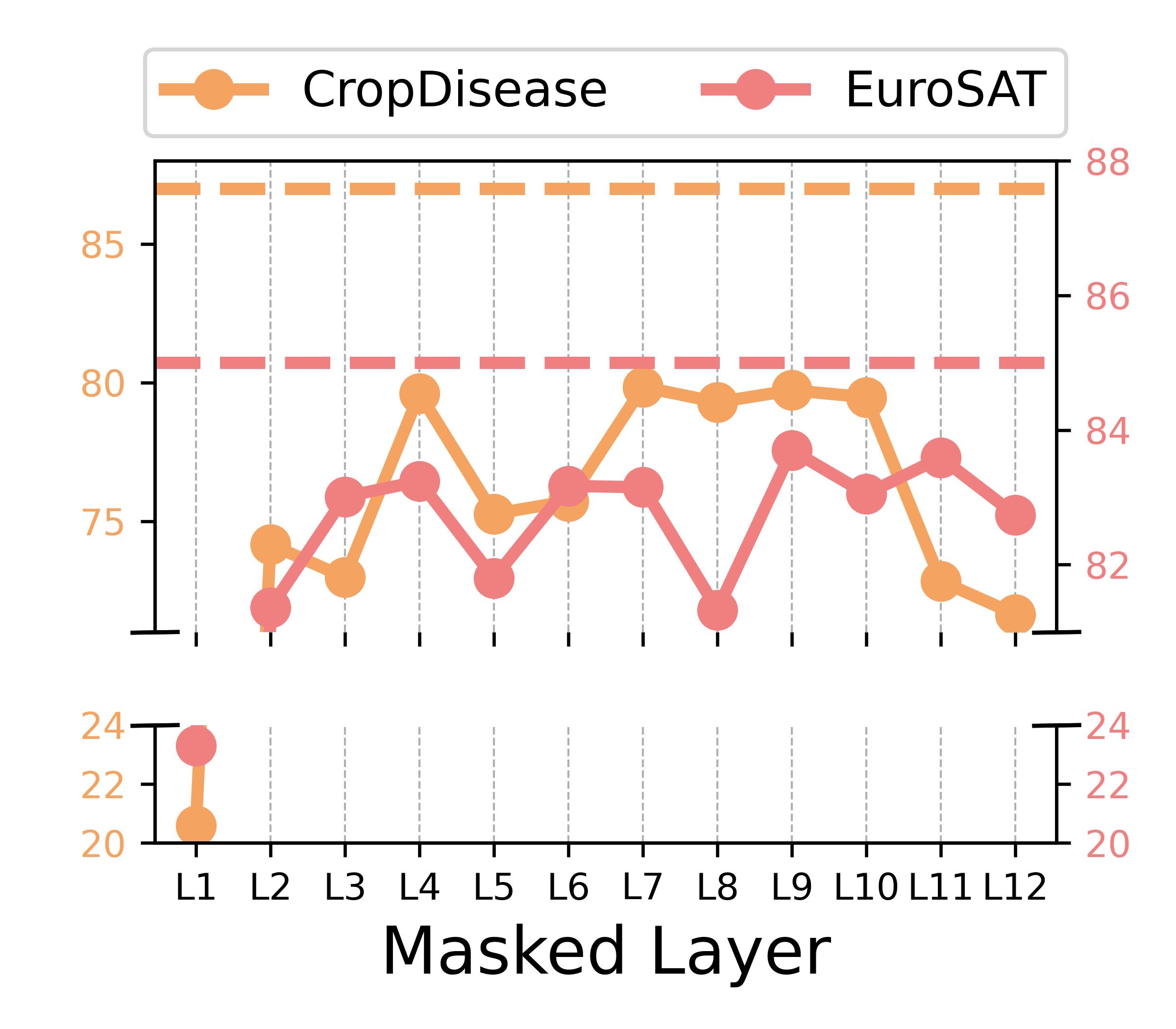}
\\
\includegraphics[width=1\linewidth]{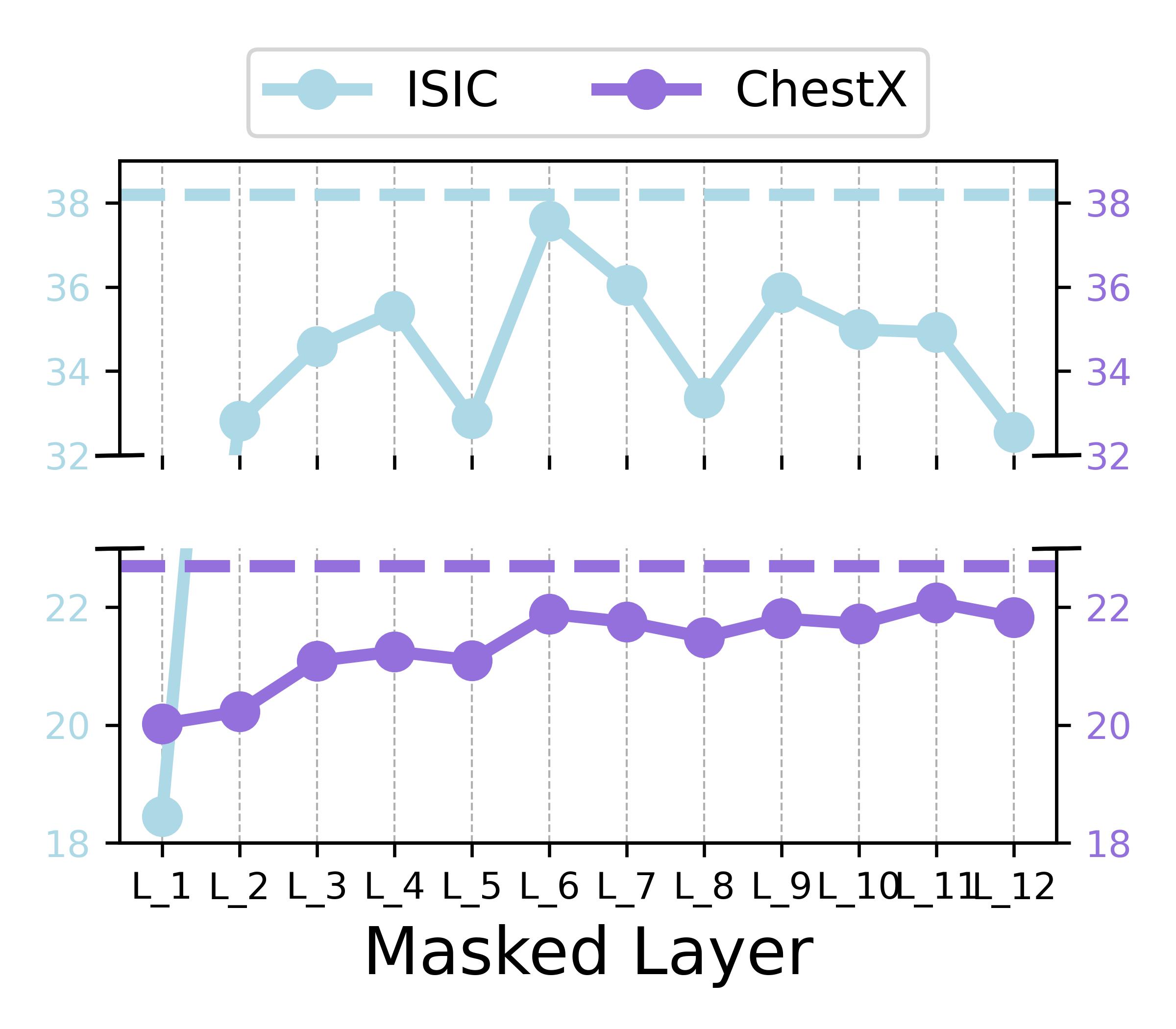}
\end{minipage}
}
\caption{(a) CLIP has two branches: a visual encoder and a text encoder. However, we find that removing certain layers of the text encoder can significantly enhance its performance in SF-CDFSL tasks. (b) Performance of 5-way 1-shot fine-tuned CLIP after removing the i-th layer (x-axis) of the text encoder. The horizontal dashed line represents the performance achieved using the full text encoder. Masking certain layers results in better performance. (c) After applying our method, the optimal performance is achieved using the full text encoder (dashed line), indicating that the lost layer no longer exists.}
\label{fig:intro}
\end{figure*}

Cross-Domain Few-Shot Learning (CDFSL)~\cite{guo2020broader,zouattention,zou2025closer} aims to address target-domain tasks (such as medical or satellite image analysis) with limited data by leveraging knowledge transferred from large-scale source-domain datasets (such as ImageNet). Traditional CDFSL works focus on the source-domain training to obtain a generalizable model for various target domains. However, in real-world scenarios, issues such as computational burden and data privacy make source-domain training infeasible. As a result, the source-free CDFSL task (SF-CDFSL)~\cite{yazdanpanah2022visual,zhuo2024prompt,xu2024step} is proposed to focus solely on the finetuning strategy on the scarce training data of target domains.

Currently, visual language models (VLMs)~\cite{radford2021learning,tschannen2025siglip,bolya2025perception}, such as CLIP~\cite{xu2024step} has achieved promising results on in-domain few-shot learning tasks. As shown in Figure~\ref{fig:intro}(left), in CLIP, there are two branches: a visual encoder and a text encoder, both composed of multi-layer networks, such as transformers. In CLIP, classification tasks are carried out by computing the similarity between image and text features. It is widely believed that CLIP contains a substantial amount of pre-trained knowledge, which enables it to adapt effectively to downstream tasks. However, when applying the CLIP model to downstream SF-CDFSL tasks, we observe an interesting phenomenon: removing a certain middle layer from the text encoder can achieve better performance than using the full text encoder, as shown in Figure~\ref{fig:intro}. We refer to these layers as \textbf{the Lost Layers}. We further discover that the lost layer phenomenon is prevalent across all backbone versions of CLIP, as shown in Figure~\ref{fig:diff_back}, and across different fine-tuning methods under the SF-CDFSL setting, as illustrated in Figure~\ref{fig:diff_peft}. In this paper, we take a step closer to interpreting this phenomenon. 

Previous work~\cite{cho2023promptstyler,schlarmann2024robust,mao2022understanding} has indicated that the information in CLIP's text encoder is more suitable for cross-domain tasks. This leads us to question whether the information in these lost layers is truly harmful for SF-CDFSL tasks. 
To investigate this, we propose two strategies to leverage the lost layer: (1) removing the lost layer during the finetuning and testing, and (2) emphasizing the lost layer by adding it to the final feature.
Interestingly, we find that manually emphasizing the lost layer can achieve the best performance. 
This means that instead of being detrimental to SF-CDFSL, \textit{the lost layers are beneficial but underutilized in the regular finetuning} of SF-CDFSL.
Then, by controlling the visual information in the classification, we finally find that it is the visual gap that prevents CLIP's visual branch from utilizing information in these layers, making these layers seem to be redundant and leading to the lost layers.
Therefore, instead of abandoning the lost layers, we need to redirect the visual branch to fully leverage the valuable information~\cite{schlarmann2024robust,mao2022understanding} in the text encoder.

To achieve this goal, we propose the VtT model to “teach the vision(V) encoder to think(t) like the text(T) encoder.” This approach aims to fully utilize the valuable pre-trained knowledge in each block of the text encoder to guide the model in solving SF-CDFSL tasks. VtT consists of three modules: the V-T Fusion module, which integrates useful information from the text into the vision features at layer-level through vision-text cross-layer scanning; the TIA module, which converts vision features into absorber tokens and inputs them into the text branch to absorb useful information at the encoder-level; and the DGSO module, which dynamically balances the trade-off between the primary classification task and the task of absorbing information from the text encoder based on gradient information. Our method ensures that information from all layers of the text encoder is utilized, thus eliminating the phenomenon of the lost layer and achieving superior performance. 

Compared with current studies~\cite{men2024shortgpt,tong2025flowcut,wang2025investigating,lad2406remarkable,gonzalez2025leveraging} on layer redundancy, our work differs in several key aspects.1) Discovery: We observe that removing the lost layer can consistently enhance performance, rather than just not significantly degrading performance (\cite{men2024shortgpt,tong2025flowcut,gonzalez2025leveraging}). Most importantly, We discoverer that \textbf{the lost layer is not redundant and can be reclaimed}. 2) Cause: We are the first to identify that the lost layer is caused by drift in the visual domain. 3) Methodology: Instead of removing layers~\cite{men2024shortgpt,tong2025flowcut,lad2406remarkable}, we reclaim the lost layer and demonstrate that this reclamation is a superior strategy compared to removal. In general, our contributions can be summarized as:

\begin{itemize}
    \item We are the first to discover that removing certain layers from the text encoder in CLIP significantly improves performance in SF-CDFSL tasks. 
    \item By extensive experiments, we find that this phenomenon arises due to changes in the visual domain, preventing some beneficial information in the text encoder from being fully utilized, leading to the lost layers.
    \item To address this issue, we propose the VtT model to “teach the vision encoder to think like the text encoder” at both the layer and encoder levels. This approach resolves the lost layer issue and improves performance.
    \item Extensive experiments on four commonly used CDFSL datasets validate the effectiveness of our method, achieving a new state-of-the-art performance.
\end{itemize}

%% file: Tex/02_related_yxz.tex
\section{Related Work}
 \textbf{Cross-Domain Few-Shot Learning (CDFSL)} aims to train a model on a source domain that can generalize effectively to target domains with limited examples. Existing methods are typically categorized into meta-learning-based approaches~\cite{fu2022wave,hu2022adversarial,wang2021cross} and transfer learning-based approaches~\cite{zhou2023revisiting,zou2024flatten,zou2025closer}. Source-Free Cross-Domain Few-Shot Learning (SF-CDFSL) introduces a stronger constraint by making source domain data inaccessible~\cite{yazdanpanah2022visual,zhuo2024prompt,xu2024step,li2025logits}. However, the influence of text layers in CLIP remains underexplored for the SF-CDFSL task.
 
\noindent\textbf{Parameter-efficient fine-tuning (PEFT)} tune the model by only a few samples, which can be grouped into prompt learning~\cite{khattak2023maple,zhu2023prompt,khattak2023self,yao2023visual,lu2023beyond,li2024promptkd}, adapters~\cite{gao2024clip,zhang2022tip,li2025logits,tang2024amu}, and LoRA (and its variants)~\cite{hu2021lora,zanella2024low}. 
However, the phenomenon of lost layers in PEFT has not been observed before.

\noindent\textbf{Layer Redundancy} is related to works~\cite{men2024shortgpt,tong2025flowcut,wang2025investigating,lad2406remarkable,gonzalez2025leveraging}, which is different from ours. 
Unlike these studies which focus on in-domain scenarios, consider these layers as redundant, and employ a removal strategy, we study the SF-CDFSL problem on VLM and find the information in these layers actually beneficial for SF-CDFSL tasks, and our method effectively re-utilizes the information. Our work provides a new perspective for analyzing similar issues. 

\begin{figure*}[!t]
\centering
\subfloat[]{
	\label{fig:diff_back}\includegraphics[width=0.26\linewidth]{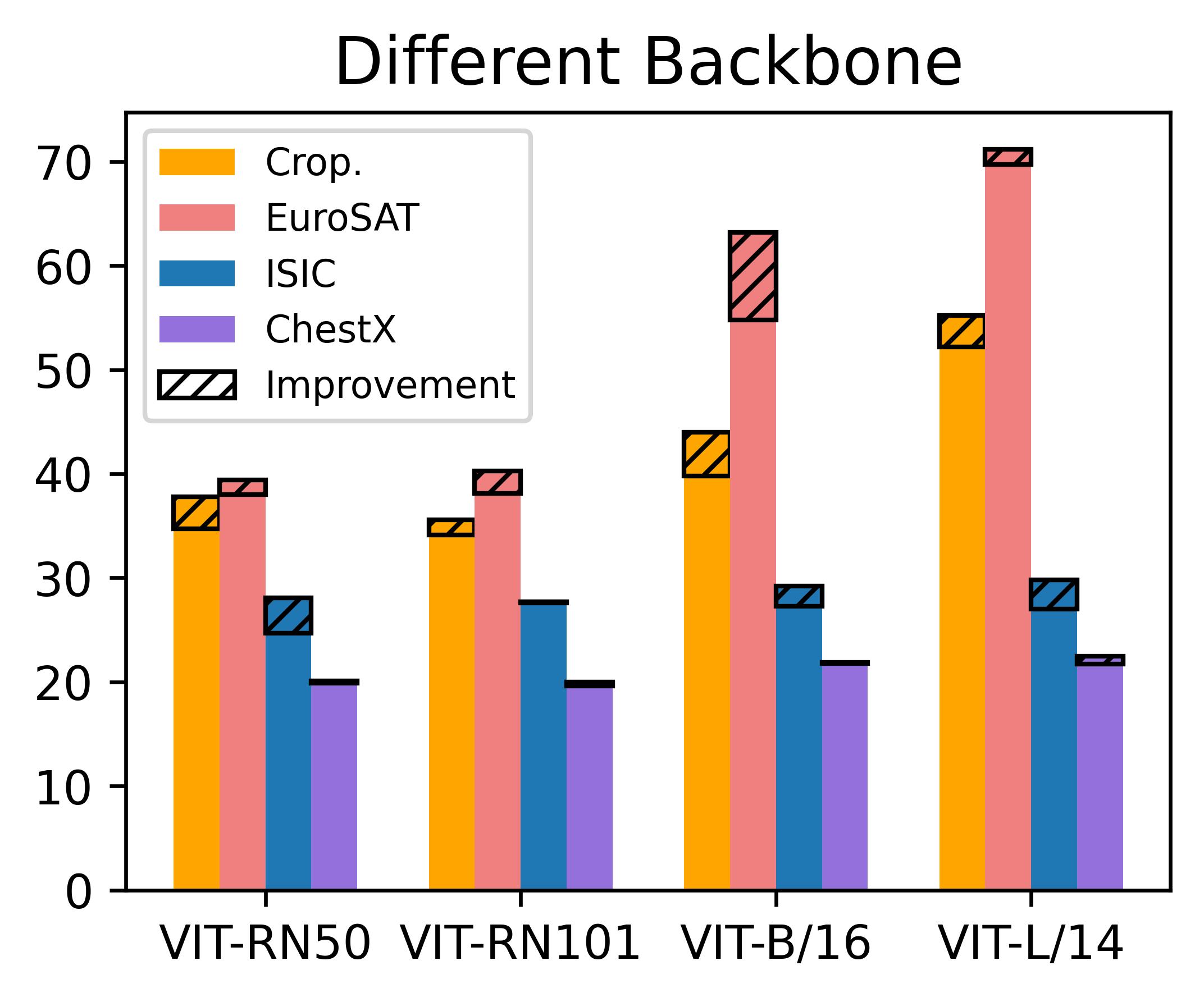}}
\subfloat[]{
	\label{fig:diff_peft}\includegraphics[width=0.26\linewidth]{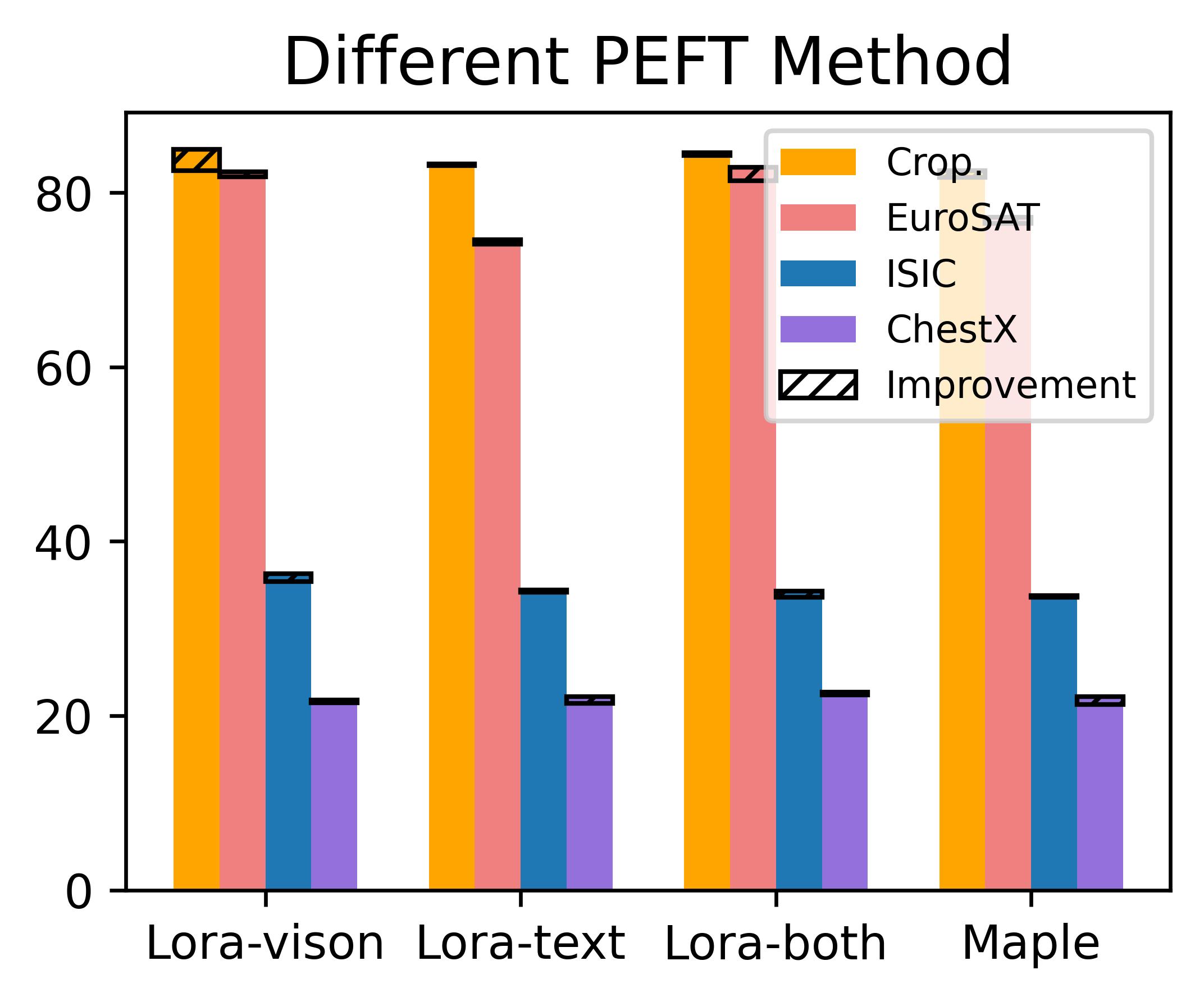}}
\subfloat[]{
	\label{fig:2_mthond}\includegraphics[width=0.20\linewidth]{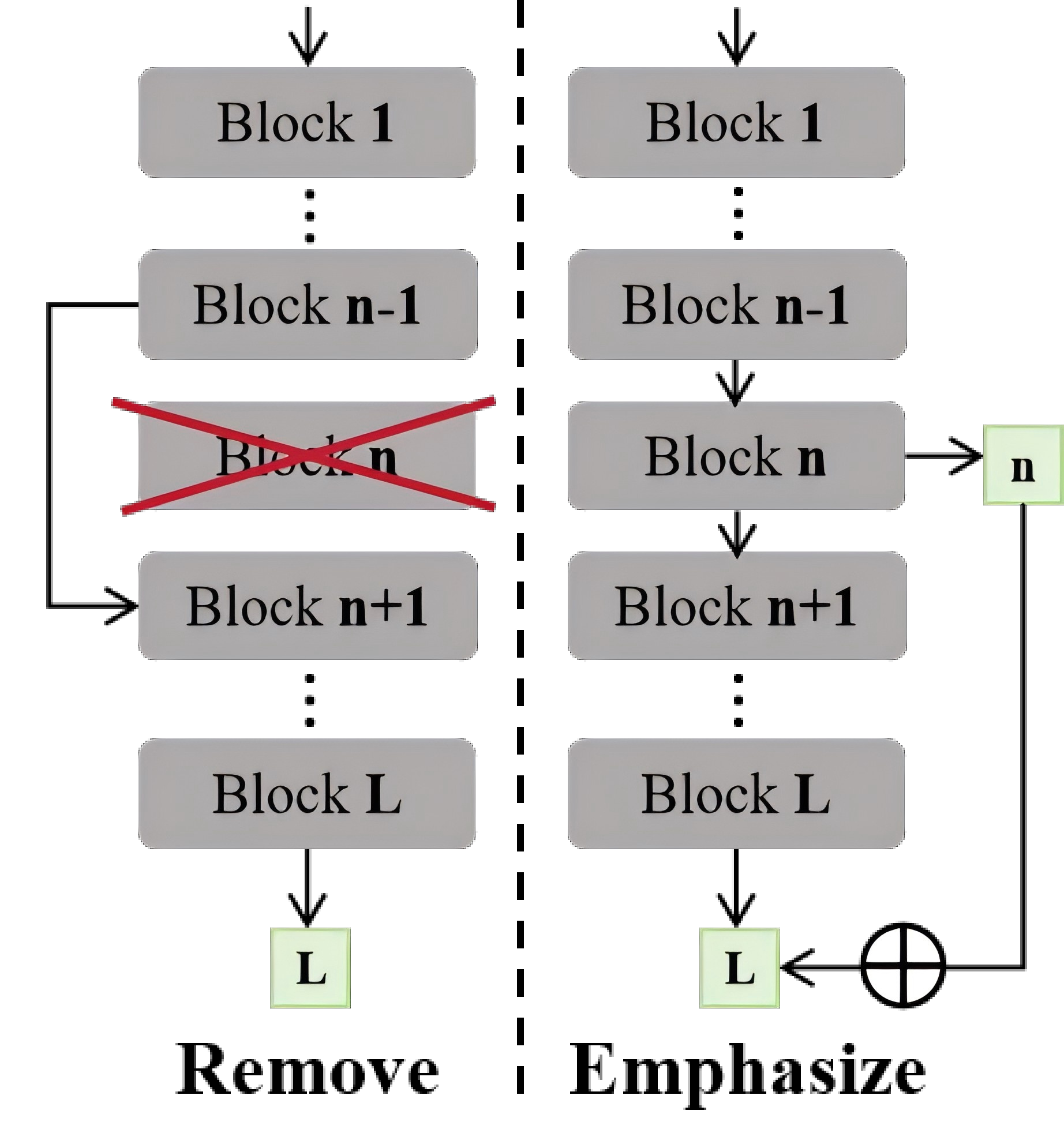}}
\subfloat[]{
	\label{fig:domain_influenc}\includegraphics[width=0.29\linewidth]{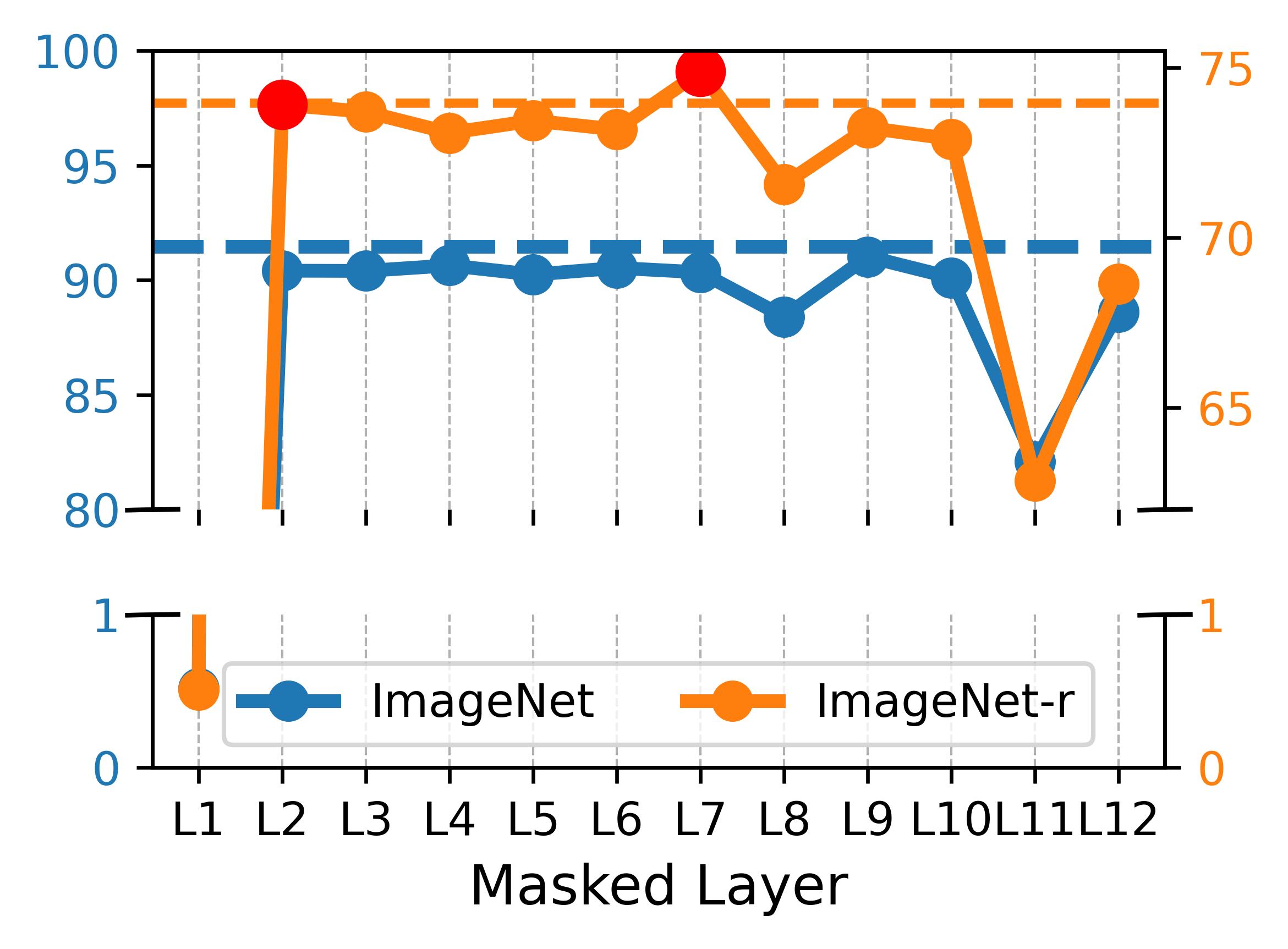}}
\caption{In SF-CDFSL tasks, (a) the lost layer is commonly present in various CLIP structures (Improvement: increased performance achieved by masking a specific layer of the text encoder), and (b) different fine-tuning methods do not effectively address this issue. Please refer to the Appendix for more detailed results. (c) Two strategies for leveraging the lost layer: Remove - eliminating the layer; Emphasize - enhancing the output of the layer using a residual approach in the final output. (d) In the source domain (ImageNet), the performance using the full text encoder (blue dashed line) is consistently optimal, thus no lost layer exists. However, after a change in the visual domain (ImageNet-R), masking the 7th layer of the text encoder significantly improves performance, indicating the reappearance of the lost layer.  }
\end{figure*}

%% file: Tex/03_preliminary_yxz.tex
\section{Preliminary}
\textbf{Source-Free Cross-Domain Few Shot Learning}:
Given a target domain dataset $D_T$, an episode $E = \{(S, Q), Y\}$ is randomly sampled. This episode is formulated as an $N$-way $K$-shot problem. Specifically, for each episode $E_i$ from $D_T$, $N$ classes are sampled with $K$ labeled images to form the support set $S$, and the same $N$ classes with $M$ different images form the query set $Q$. The label set for these $N$ classes is denoted as $Y = \{ c_i \}_{i=1}^N$. The support set $S$ is used for training, while the query set $Q$ is used to evaluate.

\noindent\textbf{Fine-tuning in CLIP}:
In classification tasks using CLIP \cite{radford2021learning}, a textual description, known as a prompt, is generated for each class, such as ``a photo of a cat". Let \( r_k \) denote the tokenized prompt for the \( k \)-th class. The language encoder \( \mathcal{F}_t \) processes \( r_k \) to produce the normalized textual embedding \( t_k = \mathcal{F}_t(r_k) \). Similarly, each image \( x_i \) is processed by the visual encoder \( \mathcal{F}_v \) to obtain the normalized visual embedding \( f_i = \mathcal{F}_v(x_i) \). Then the classification probability is calculated as: 
\begin{equation}\label{eq:cross_sim_and_p}
\setlength{\abovedisplayskip}{5pt}
\setlength{\belowdisplayskip}{5pt}
s_{i, j} = Sim(f_i,t_j), \quad  p_{i, j} = \frac{\exp\left(s_{i, j} / \tau\right)}{\sum_{j=1}^K \exp\left(s_{i, j} / \tau\right)},
\end{equation}
where $s_{i, j}$ represents the cosine similarity between image sample $i$ and text prompt $j$, and $\tau$ is the temperature coefficient. In the CLIP model, the value of $\tau$ is set to 0.01.

Many works explored fine-tuning techniques in CLIP~\cite{khattak2023maple,zanella2024low,xu2024step,huang2024lp++,li2025logits} for in-domain few-shot learning. Despite differences in learning strategies and parameter update mechanisms, most methods fine-tune the model by minimizing the cross-entropy loss:
\begin{equation}\label{eq:cross_loss}
\setlength{\abovedisplayskip}{5pt}
\setlength{\belowdisplayskip}{5pt}
L_{\mathrm{ce}}=-\frac{1}{N} \sum_i \log \frac{\exp \left(s_{i, i} / \tau\right)}{\sum_j \exp \left(s_{i, j} / \tau\right)}.
\end{equation}

%% file: Tex/04_ana_yxz.tex
\section{Analysis on Lost Layers in CLIP}


\begin{table}[!t]

\caption{Performance after 5-way 1-shot fine-tuning using two strategies, showing both strategies improve the performance but emphasizing these layers is better, implying useful but underutilized information in the lost layers. However for both strategies, wrong layer selection (wls) can cause a large performance drop}
\centering
    \begin{adjustbox}{max width=0.95\linewidth}
    \begin{tabular}{p{0.30\linewidth}cccccc}
    \toprule
    Method & Crop. & EuroSAT & ISIC & ChestX & Avg\\
    \midrule
    \raggedright Removal* & 85.1  &82.2  & 36.4 & 21.6  & 56.3\\
    \midrule
    \raggedright Removal (wls)* & 72.6  &79.8  & 30.6 & 20.1  & 50.7\\
    \midrule
    \raggedright Emphasis* & 85.3  &82.7  & 37.0 & 22.0  & 56.7\\
    \midrule
    \raggedright Emphasis (wls)* & 81.9  &80.9  & 32.7 & 21.0  & 54.1\\
    \midrule
    \raggedright \textbf{VtT(OURS)} & \textbf{87.0}  &\textbf{85.0}  & \textbf{38.2} & \textbf{22.7}  & \textbf{58.2}\\
    \bottomrule
    \end{tabular}
    \end{adjustbox} 
\label{tab:2method}
\end{table}

\subsection{Truly Redundant or Underutilized?}
The results in Figure~\ref{fig:diff_back} and~\ref{fig:diff_peft} indicate that some layers in the text encoder appear to be redundant. However, previous studies~\cite{schlarmann2024robust} indicate that the knowledge within CLIP's text encoder is more suitable for cross-domain tasks. This leads us to question whether the information in these lost layers is truly redundant. To explore this, we design two strategies to utilize the lost layers based on two assumptions:

\noindent\textbf{(1) If such a layer is indeed redundant}, we can simply remove it (Remove), as shown in Figure~\ref{fig:2_mthond}(left):
\begin{equation}
\setlength{\abovedisplayskip}{5pt}
\setlength{\belowdisplayskip}{5pt}
\label{eq:remove}
t^{-i}_{k} = \mathcal{F}^{-i}_t(r_k), 
\end{equation}
where $\mathcal{F}^{-i}_t$ denote the text encoder with the $i$-th layer removed, and let $t^{-i}_{k}$ represent the text feature extracted by this text encoder for the prompt $r_k$.

\noindent\textbf{(2) If this layer is beneficial but underutilized}, we can enhance its contribution to the final output to better utilize this information (Emphasize), as illustrated in Figure~\ref{fig:2_mthond}(right):
\begin{equation}
\setlength{\abovedisplayskip}{5pt}
\setlength{\belowdisplayskip}{5pt}
\label{eq:emphas}
t^{+i}_{k} = (1 - \gamma)t_{k} + \gamma t^{i}_{k}, 
\end{equation} 
where $t^{i}_{k}$ represents the output EOS token of the $i$-th layer of the text encoder corresponding to $t_{k}$. The parameter $\gamma$ is used to control the weight of the reinforcement applied to $t^{i}_{k}$. In all experiments, we set $\gamma = 0.2$.

When using the two strategies, we apply the same value of $i$, meaning we perform operations on the same layer. We replace $t_{k}$ in Equation~\ref{eq:cross_sim_and_p} with $t^{-i}_{k}$ and $t^{+i}_{k}$, respectively, and then train the model according to Equation~\ref{eq:cross_loss}. 
The results are shown in Table~\ref{tab:2method}, which indicate that the Emphasize strategy significantly enhances model performance and outperforms the Remove strategy. This finding supports assumption (2), demonstrating that the information in these layers is indeed beneficial for SF-CDFSL tasks. However, it is worth noting that even the Remove strategy leads to performance improvement. 
This is why we refer to these layers as “the lost layers”: under the SF-CDFSL setting, ignoring the information in these layers can improve performance. However, this is not the optimal choice, as the text encoder contains substantial pre-trained, domain-independent knowledge. The information in these layers is beneficial for SF-CDFSL tasks but remains in a lost state. A better strategy is to effectively re-utilize this information. 

Note that Removal and Emphasis require a per dataset layer search in which a single layer is masked or focused to identify the best layer index. This procedure is unreliable for addressing the lost-layer problem because the optimal index i varies across datasets (e.g., 6, 4, 6, 4 for Removal), and wrong layer selection (wls) can cause a large performance drop. In Table~\ref{tab:2method}, Removal and Emphasis are marked with an asterisk as “oracle” because the best-layer results require extra layer search and validation. Even under this idealized setting, our method still performs best.

\subsection{What Causes the Lost Layer?}
We analyze whether visual or text factors are responsible for the lost layer. Unlike classification tasks on natural domain datasets like ImageNet~\cite{deng2009imagenet}, SF-CDFSL tasks primarily vary in image styles or domain information. To show that changes in the visual domain cause the lost layer phenomenon, we alter the visual domain while keeping semantic information constant and observe the lost layer occurrence. We conduct classification experiments using CLIP on the ImageNet and ImageNet-R datasets (a cross-domain version of ImageNet), ensuring consistent category information across 200 overlapping categories. The results (Figure~\ref{fig:domain_influenc}) show that the lost layer phenomenon is absent in the original domain (ImageNet) but appears in cross-domain scenarios (ImageNet-R), even with unchanged category information. We also verify that semantic information is not the cause of the lost layer, see results in the Appendix.


\subsection{Key Insights}
The previous analysis indicates that:
(1) Information within the lost layers is inherently beneficial for SF-CDFSL tasks but has not been utilized, instead of simply being harmful or redundant as in \cite{men2024shortgpt,tong2025flowcut,wang2025investigating,lad2406remarkable,gonzalez2025leveraging}. (2) The lost layer is caused by changes in the visual domain, suggesting that issues within the visual branch lead to the neglect of information in the text encoder under SF-CDFSL scenarios. 
Therefore, it is essential to redirect the visual branch to fully leverage the valuable information in the text encoder to address SF-CDFSL tasks effectively.

%% file: Tex/05_method_yxz.tex
\section{Method}
\begin{figure*}[t]
\centering
\includegraphics[width=0.95 \linewidth]{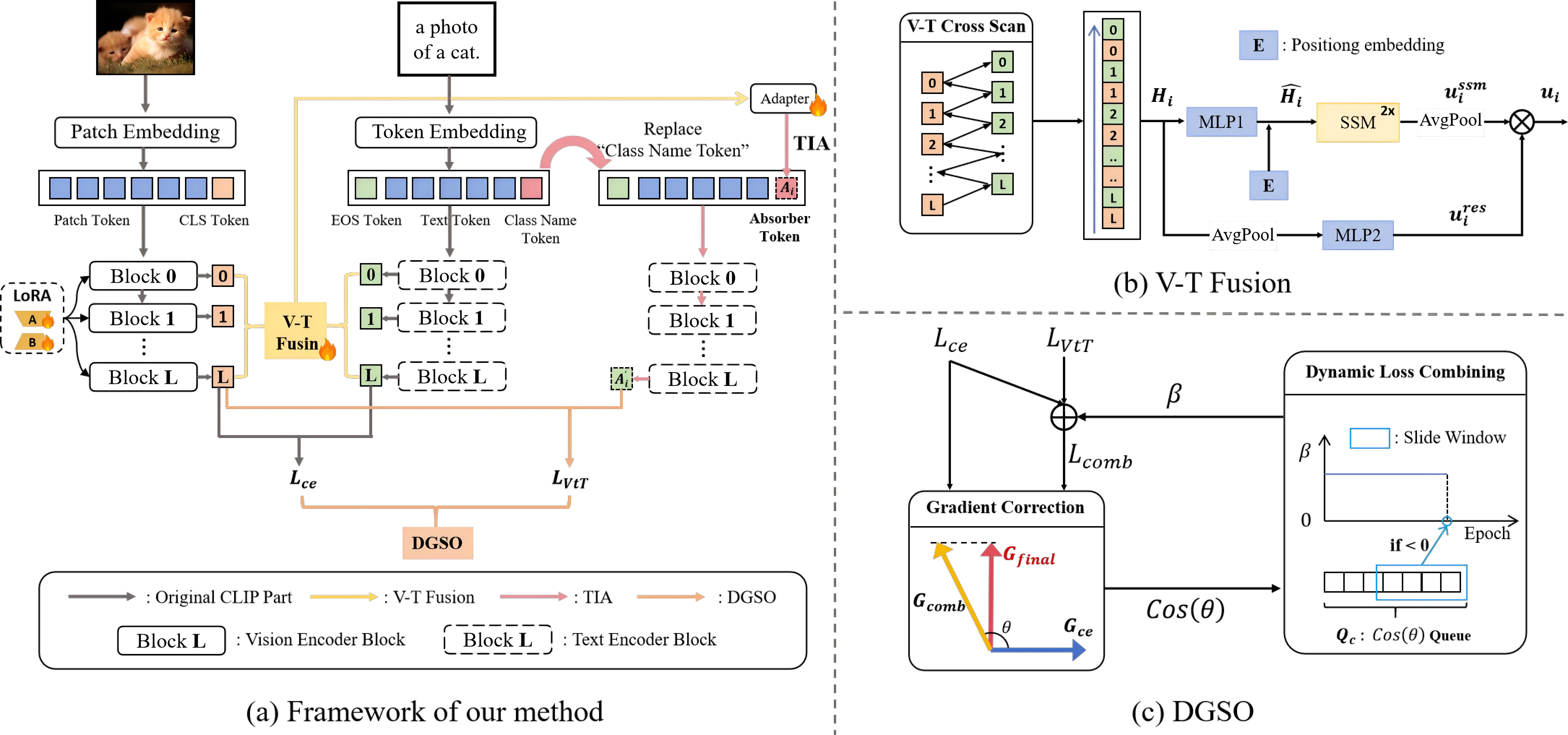}
\caption{(a) The overall architecture of the VtT model. First, the V-T Fusion module integrates visual and textual features at the layer level (yellow lines). Then, the TIA module absorbs information from the text encoder at the encoder level (pink lines). Finally, the DGSO module optimizes the model using the outputs from the previous modules and gradient information (orange lines). (b) The V-T Fusion module interleaves the outputs of the visual and text encoders from deep to shallow layers and integrates them using the SSM network. (c) The DGSO module removes gradients that conflict with the main task (classification) before optimizing the model. It also determines when to stop using the VtT model based on the extent of gradient conflicts.}
\label{fig:method}
\end{figure*}

Our analysis has shown that changes in the visual domain result in the under-utilization of useful information within the text encoder. Therefore, we need to re-utilize the effective information in each block of the text encoder to better address SF-CDFSL tasks. To address this issue, we propose the TtV model, as depicted in Figure~\ref{fig:method}. 

Our model first employs the V-T Fusion module to integrate useful information from the text into the vision features at the layer level (indicated by the yellow line). After this initial fusion, the TIA module constructs absorber tokens and inputs them into the text encoder to capture useful information at the encoder level (indicated by the pink line). The final fusion output, now enriched with the effective information from the text branch, are treated as expert information. Finally, the DGSO module uses this expert information along with gradient information to guide the model's learning process (indicated by the orange line).

\subsection{Visual-Text Layer-Level Fusion (V-T Fusion)}
In the first stage (indicated by the yellow line), we integrate information from both branches at the layer level. Inspired by Mamba-related techniques~\cite{li2024mamba,gu2023mamba}, we employ the concept of sequential modeling to systematically fuse visual and text features at each layer. 

Firstly, we propose visual-text cross-scaning to combine the layer-wise outputs of the text and vision branches, as shown in Figure~\ref{fig:method}b. We reshape the output CLS/EOS token of each layer from both text encoder and vision encoder into a sequence $H_i \in \mathbb{R}^{2l}$ for sample $i$, with a scanning direction that alternates from deep to shallow layers (Refer to Ablation Study for discussion of the scanning direction):
\begin{equation}
\setlength{\abovedisplayskip}{5pt}
\setlength{\belowdisplayskip}{5pt}
\label{eq:cross_scan}H_{i}=(f_i^l, t_i^l, f_i^{l-1}, t_i^{l-1}, \cdots , f_i^1, t_i^1),
\end{equation}
here, $l$ represents the total number of layers in the encoder, $f_i^l$ denotes the CLS token obtained from the $l$-th layer of the vision encoder for sample $i$. Similarly, $t_i^l$ represents the EOS token obtained from the $l$-th layer of the text encoder for the prompt corresponding to the category of sample $i$.

We then utilize a State Space Model (SSM)~\cite{gu2021efficiently} to aggregate the visual and textual information in $H_i$. This aggregation process comprises a residual branch and an SSM branch. In the residual branch, $H_i$ undergoes averaging and is then processed through an MLP layer: 
\begin{equation} 
\setlength{\abovedisplayskip}{5pt}
\setlength{\belowdisplayskip}{5pt}
{\mu}_{i}^{\text {res }}=M_2(\operatorname{AvgPool}(H_{i})) , 
\end{equation}
where AvgPool denotes the average pooling function, and $M_2$ is an MLP layer. For the SSM branch, $H_i$ is initially processed by an MLP layer $M_1$ and then combined with the position embedding as: 
\begin{equation} 
\setlength{\abovedisplayskip}{5pt}
\setlength{\belowdisplayskip}{5pt}
\hat{H_i}=M_2(H_{i})+E, \end{equation} 
and $\hat{H_i}$ is subsequently processed by a 2-layer SSM, followed by averaging: 
\begin{equation} \label{eq:ssm}
\setlength{\abovedisplayskip}{5pt}
\setlength{\belowdisplayskip}{5pt}
{\mu}_{i}^{\text {ssm }}=\operatorname{AvgPool}\left(\operatorname{SSM}\left(\hat{H_i}\right)\right) .
\end{equation}

The final output is obtained by summing the outputs of the residual branch and the SSM branch: 
\begin{equation} 
\setlength{\abovedisplayskip}{3pt}
\setlength{\belowdisplayskip}{3pt}
{\mu}_i={\mu}_{i}^{\text {res }}+{\mu}_{i}^{\text {ssm}}. 
\end{equation}

\subsection{Text Encoder Information Absorption (TIA)}

At the encoder level, inspired by modality conversion methods~\cite{mistretta2025cross}, we input visual features into the text encoder to absorb the knowledge missing from the visual branch. Specifically, we input the layer-fused ${\mu}_i$
into the text branch to further absorb encoder-level information, as shown in Figure~\ref{fig:method}a, indicated by the pink line.

Firstly, ${\mu}_i$ is subjected to a learnable adapter to map it into the text feature space, which aims to transform ${\mu}_i$ into a token understandable by the text branch: 
\begin{equation} 
\setlength{\abovedisplayskip}{5pt}
\setlength{\belowdisplayskip}{5pt}
A_i =\operatorname{ReLU}\left(\operatorname{M_3}\left({\mu}_i\right)\right), 
\end{equation} 
where $\operatorname{ReLU}$ is the activation function, and $M_3$ is a linear layer. The resulting $A_i$ is referred to as the absorb token, as it will be input into the text branch to absorb knowledge.

For a template such as “a photo of a \{\}”, the tokenized prompt $r_i$ corresponding to the category of sample $i$ is: 
\begin{equation} 
\setlength{\abovedisplayskip}{6pt}
\setlength{\belowdisplayskip}{6pt}
r_i = [\mathrm{a}]_1[\mathrm{photo}]_2 [\mathrm{of}]_3 [\mathrm{a}]_4 [\mathrm{CLASS}]_5, 
\end{equation} 
where $[\mathrm{a}]_1$ is the token vector for “a”. We replace the token of to the class name, $[\mathrm{CLASS}]_5$, with $A_i$ to generate $r_{i}^{\prime}$:
\begin{equation} 
\setlength{\abovedisplayskip}{6pt}
\setlength{\belowdisplayskip}{6pt}
\label{eq:gen_r}r_{i}^{\prime} = [\mathrm{a}]_1[\mathrm{photo}]_2 [\mathrm{of}]_3 [\mathrm{a}]_4 [{A_i}]_5. 
\end{equation} 
Next, $r_{i}^{\prime}$ is input into the text encoder $\mathcal{F}_t$ to absorbe information from the text encoder:
\begin{equation}
\setlength{\abovedisplayskip}{3pt}
\setlength{\belowdisplayskip}{3pt}
\label{eq:gen_a}A_{i}^{\prime} = \mathcal{F}_t(r_{i}^{\prime}),
\end{equation}
$A_{i}^{\prime}$ combines the layer-level detailed knowledge from both the visual and text branches and the encoder-level holistic knowledge from the text branch. Then,the $L_{TtV}$
loss is employed to guide the learning process using $A_{i}^{\prime}$:
\begin{equation}
\setlength{\abovedisplayskip}{3pt}
\setlength{\belowdisplayskip}{3pt}
\label{eq:vtt_loss}
L_{\text {VtT}}=-\frac{1}{N} \sum_{i=1}^{N}\frac{A_{i}^{\prime}}{\left\|A_{i}^{\prime}\right\|_2} \cdot \frac{f_i}{\left\|f_i\right\|_2}.
\end{equation}
The objective of $L_{\text{VtT}}$ is to ensure that the visual feature $f_i$ of sample $i$ closely matches $A_{i}^{\prime}$. This approach serves two main purposes: 1) It distills the useful textual information contained in $A_{i}^{\prime}$ into the visual features, addressing the issue of underutilized textual information. 2) It aligns the visual features $f_i$ with the output $A_{i}^{\prime}$ from the text branch, thus facilitating the alignment between textual and visual features.





\subsection{\mbox{Dynamic Gradient Supervised Optimization}}\label{sec:DGSO}
Our model's learning process consists of two tasks: the first is the CLIP classification task, represented by $L_{ce}$ (Eq.~\ref{eq:cross_loss}); the second task corresponds to $L_{VtT}$ (Eq.~\ref{eq:vtt_loss}), which aims to enhance the utilization of information from the text encoder. Therefore, the combined learning objective is 
\begin{equation} 
\setlength{\abovedisplayskip}{3pt}
\setlength{\belowdisplayskip}{3pt}
L_{comb} = L_{ce} + \beta L_{VtT}, 
\end{equation} 
where $\beta$ is a hyperparameter.

\begin{table*}[t]
\belowrulesep=0pt
\aboverulesep=0pt
\caption{
The accuracy(\%) of four target domain datasets under 5-way 1-shot and 5-way 5-shot tasks. See extended results in Appendix.}
\label{tab:exp_cdfsl}    
\centering
\begin{adjustbox}{max width=0.95\linewidth}
\begin{tabular}{*{8}{c|c c|c c c c c|} }  
\toprule 
 \multirow{1}*{Task}  &  \multirow{1}*{Method}  &\multirow{1}*{backbone} & \multirow{1}*{ISIC} & \multirow{1}*{EuroSAT} & \multirow{1}*{CropDisease} & \multirow{1}*{ChestX} & \multirow{1}*{Avg} \\           
 \midrule
  \multirow{14}{*}{\rotatebox{90}{5-way 1-shot}} 
  &StepSTP~\cite{xu2024step} &ViT/CLIP &32.97±0.27 &70.01±0.21 &84.84±0.72 &\textbf{22.84±0.95} &52.68 \\
  &CoOp~\cite{zhou2022learning}  &ViT/CLIP  &32.86±0.47 &72.08±0.66 &80.50±0.74 &21.65±0.32 &51.77 \\
  &Tip-Adapter~\cite{zhang2021tip}  &ViT/CLIP &32.68±0.37 &75.44±0.51 &77.15±0.66 &22.24±0.26 &51.87 \\
  &AMU-Tuning~\cite{tang2024amu}  &ViT/CLIP &32.29±0.67 &72.24±0.71 &80.20±0.86 &21.56±0.36 &51.57 \\
  &LP++~\cite{huang2024lp++}  &ViT/CLIP &33.63±0.41 &73.05±0.55 &81.84±0.66 &21.72±0.42 &52.56 \\
  &LDC~\cite{li2025logits}  &ViT/CLIP &33.72±0.46 &74.19±0.52 &83.77±0.81 &22.12±0.36 &53.45 \\
  &Maple~\cite{khattak2023maple}  &ViT/CLIP &33.38±0.49 &76.05±0.63 &81.78±0.72 &21.09±0.31 &53.07 \\
  &\textbf{Maple + VtT (OURS)}  &ViT/CLIP &34.06±0.53 &82.25±0.75 &82.66±0.72 &21.64±0.34 &55.15 \\
  &CLIP-LoRA-Vision~\cite{zanella2024low}  &ViT/CLIP &36.40±0.42 &81.72±0.52 &84.22±0.62 &21.86±0.32 &55.97 \\
  \rowcolor{cyan!10} \cellcolor{white}
  &\textbf{CLIP-LoRA-Vision + VtT (OURS)}  &ViT/CLIP &\textbf{38.20±0.45} &\textbf{85.01±0.41} &\textbf{87.00±0.53} &22.70±0.33 &\textbf{58.23} \\
  \cline{2-8}
  &SigLIP2-LoRA~\cite{tschannen2025siglip} &ViT/SigLip2  &33.47 &74.16 & 87.50 &21.44 & 54.14 \\
  \rowcolor{cyan!10} \cellcolor{white}
  &\textbf{SigLIP2-LoRA + VtT (OURS)} &ViT/SigLip2 &\textbf{35.34} &\textbf{76.10} & \textbf{89.72} &\textbf{22.00} & \textbf{55.79} \\
  \cline{2-8}
  &PE-Core-LoRA~\cite{bolya2025perception} &ViT/PE-Core  &40.89 &84.49 & 91.75 &22.02 & 59.78 \\
  \rowcolor{cyan!10} \cellcolor{white}
  &\textbf{PE-Core-LoRA + VtT (OURS)} &ViT/PE-Core &\textbf{42.20} &\textbf{86.16} & \textbf{92.61} &\textbf{23.04} & \textbf{61.00} \\
  \midrule
  \midrule
   \multirow{14}{*}{\rotatebox{90}{5-way 5-shot}} 
  &StepSTP~\cite{xu2024step} &ViT/CLIP &52.12±0.36 &89.40±1.05 &96.01±0.88 &26.36±0.97 &65.97 \\
  &CoOp~\cite{zhou2022learning}  &ViT/CLIP &45.78±0.75 &85.88±0.49 &93.31±0.57 &23.35±0.50 &62.08 \\
  &Tip-Adapter~\cite{zhang2021tip}  &ViT/CLIP &46.96±0.59 &87.24±0.33 &94.19±0.39 &24.07±0.44 &63.12 \\
  &AMU-Tuning~\cite{tang2024amu}  &ViT/CLIP &44.60±0.62 &88.47±0.39 &94.26±0.52 &23.34±0.41 &62.66 \\
  &LP++~\cite{huang2024lp++}  &ViT/CLIP &48.49±0.44 &87.48±0.42 &94.47±0.38 &23.89±0.29 &63.58 \\
  &LDC~\cite{li2025logits}  &ViT/CLIP  &49.70±0.33 &90.82±0.22 &96.71±0.34 &25.89±0.21 &65.78 \\
  &Maple~\cite{khattak2023maple}  &ViT/CLIP &48.35±0.75 &89.04±0.52 &93.50±0.54 &22.96±0.50 &63.46 \\
  &\textbf{Maple + VtT (OURS)}  &ViT/CLIP &49.81±0.78 &92.24±0.42 &94.62±0.53 &24.04±0.50 &65.18 \\
  &CLIP-LoRA-Vision~\cite{zanella2024low}  &ViT/CLIP &52.22±0.71 &93.31±0.47 &95.88±0.42 &24.61±0.47 &66.50 \\
  \rowcolor{cyan!10} \cellcolor{white}
  &\textbf{CLIP-LoRA-Vision + VtT (OURS)}  &ViT/CLIP &\textbf{56.20±0.41} &\textbf{94.58±0.31} &\textbf{97.21±0.35} &\textbf{26.42±0.31} &\textbf{68.57} \\
  \cline{2-8}
  &SigLIP2-LoRA~\cite{tschannen2025siglip} &ViT/SigLip2 &51.79 &91.39 & 96.43
  &24.24 & 65.96 \\
  \rowcolor{cyan!10} \cellcolor{white}
  &\textbf{SigLIP2-LoRA + VtT (OURS)} &ViT/SigLip2 &\textbf{55.11} &\textbf{92.70} & \textbf{97.63} &\textbf{25.54} & \textbf{67.75} \\
  \cline{2-8}
  &PE-Core-LoRA~\cite{bolya2025perception} &ViT/PE-Core &58.81 &94.07 & 97.25 &24.44 & 68.64 \\
  \rowcolor{cyan!10} \cellcolor{white}
  &\textbf{PE-Core-LoRA + VtT (OURS)} &ViT/PE-Core &\textbf{60.03} &\textbf{94.67} & \textbf{98.36} &\textbf{27.05} & \textbf{70.05} \\
\bottomrule
\end{tabular}
\end{adjustbox}
\end{table*}

\subsubsection{Gradient Correction}
Inspired by the success of knowledge distillation~\cite{zhu2023prompt} , we treat the classification task as the primary task ($L_{ce}$) and compare the optimization direction of $L_{comb}$ with the optimization direction of the primary task ($L_{ce}$) to regularize the final parameter optimization direction, as shown in Figure~\ref{fig:method}c. Specifically, we calculate the cosine similarity between the optimization directions of the model parameters $\theta$ under the two loss functions $L_{ce}$ and $L_{comb}$:
\begin{equation}
\setlength{\abovedisplayskip}{6pt}
\setlength{\belowdisplayskip}{6pt}
C_{\theta}=\frac{\nabla_\theta L_{\text {ce }}(x)^T}{\left\|\nabla_\theta L_{\text {ce}}(x)\right\|_2} \frac{\nabla_\theta L_{\text {comb}}(x)}{\left\|\nabla_\theta L_{\text {comb}}(x)\right\|_2}.
\end{equation}
We denote the gradient directions of $L_{ce}$ and $L_{comb}$ as $G_{ce}=\nabla_{\theta} L_{ce}(x)$ and $G_{comb}=\nabla_{\theta} L_{comb}(x)$, respectively. The relationships between $G_{ce}$ and $G_{comb}$ are two-fold: (1) $C_{\theta} > 0$, which indicates that the optimization direction of $L_{comb}$ does not conflict with the main task $L_{ce}$. In this case, we can safely set the updated gradient direction $G_{final}$ as $G_{comb}$. (2) $C_{\theta} < 0$, which indicates that $L_{comb}$ conflicts with the main task. In other words, optimizing the parameters $\theta$ following $G_{comb}$ would be detrimental to the primary task, i.e., the classification task. In this scenario, we perform gradient correction, which project $G_{comb}$ onto the orthogonal direction of $G_{ce}$ to optimize the model, thereby avoiding an increase in $L_{ce}$, as shown in Figure~\ref{fig:method}c. The optimization strategy is mathematically formulated as:
\begin{equation}
\setlength{\abovedisplayskip}{6pt}
\setlength{\belowdisplayskip}{6pt}
G_{\text {final}}= \begin{cases}G_{comb}, & \text { if } C_{\theta} \geq 0 \\ 
G_{comb}-\frac{\left\|G_{\mathrm{comb}}\right\|_2}{\left\|G_{ce}\right\|_2} C_{\theta}G_{ce}, & \text { otherwise. }\end{cases}
\end{equation}
Here, \( G_{\text{final}} \) has the conflicting gradients subtracted.

\subsubsection{Dynamic Loss Combining}
During the fine-tuning, $C=\frac{1}{\left|\mathcal{F}_v\right|} \sum_\theta^{\mathcal{F}_v} C_{\theta} $ can measure the impact of our proposed $L_{VtT}$ loss during training. When $C$ is generally positive, the strategy to “teach the visual encoder to think like the text encoder” is beneficial for the primary task (classification). Conversely, when $C$ consistently becomes negative, it indicates that the extraction of information from the text encoder has become sufficient and begins to dominate, which is detrimental to the primary task. Consequently, we dynamically decide when to use and when to stop using $L_{VtT}$ by maintaining a queue of $C$. This design aligns with some optimization-related strategies~\cite{bai2021understanding}, such as early stopping. Firstly, we maintain a queue $Q_C$:
\begin{equation}
\label{eq:que}
\setlength{\abovedisplayskip}{5pt}
\setlength{\belowdisplayskip}{5pt}
Q_C=(C^1,C^2,C^3,\cdots,C^{e-1},C^e)
\end{equation}
where $C^e$ denotes the value of $C$ at the $e$-th epoch of model training. At each epoch, for instance, the $e$-th epoch, we use a sliding window of length $\lambda$ to compute the average of $Q_C$ from epoch $(e-\lambda)$ to the current epoch:
\begin{equation}
\setlength{\abovedisplayskip}{5pt}
\setlength{\belowdisplayskip}{5pt}
\label{eq:slide}
M_e= \begin{cases}Mean(C^1,C^2,\cdots,C^e), & \text { if } e \leq \lambda \\ 
Mean(C^{e-\lambda},C^2,\cdots,C^e), & \text { otherwise. }\end{cases}
\end{equation}
Based on $M_e$, we can determine when to stop using $L_{VtT}$, and once $L_{VtT}$ is stopped, it will not be reactivated:
\begin{equation}
\setlength{\abovedisplayskip}{5pt}
\setlength{\belowdisplayskip}{5pt}
\label{eq:comb}
L_{comb}= \begin{cases}L_{ce} + \beta L_{VtT}, & \text { if } M_e \ge 0 \\ 
L_{ce}, & \text { otherwise. }\end{cases}
\end{equation}

\subsection{Stable Training and Efficient Inference}
Our VtT model is an easily integrable plugin that does not require any modifications to the backbone network structure. Its primary function is to assist in the fine-tuning process of the CLIP model for SF-CDFSL tasks. 

To ensure stable training of the VtT module: 1) The V-T cross scan (Equation~\ref{eq:cross_scan})integrates information from deep to shallow layers, ending with the output of the first text layer. This ensures that, in the early stages of VtT module training, the network's output matches the first text layer's output. At this stage, the TIA process functions as the initial forward pass of the text branch ($A_{i}^{\prime} \approx t_i$), ensuring stability.
2) Following~\cite{xu2024step}, data augmentation is applied.
3) The B matrix in the LoRA component is initialized to zero~\cite{hu2021lora}.

After fine-tuning, all VtT-related parameters are removed and the original CLIP prediction method is used for classification. Specifically, similarity is calculated using Equation~\ref{eq:cross_sim_and_p}, followed by classification. Thus, our method does not introduce any additional inference overhead.

%% file: Tex/06_exp_yxz.tex
\section{Experiments}

\subsection{Datasets and Implementation Details}
\noindent
\textbf{Datasets.}
Following previous works~\cite{yazdanpanah2022visual,zhuo2024prompt,xu2024step}, we do not utilize source domain datasets and finetune our model directly on the target domain. For the target domain we utilize CropDisease~\cite{mohanty2016using}, EuroSAT~\cite{helber2019eurosat}, ISIC~\cite{codella2019skin}, and ChestX~\cite{wang2017chestx}, which are cross-domain datasets from the domain of agriculture, remote sensing, and medical data with significant domain gaps.

\noindent
\textbf{Implementation Details.}
We adopt the ViT-Base/16 network as the backbone, utilizing parameters pre-trained by CLIP~\cite{radford2021learning}. In every layer of the visual branch, we incorporate the LoRA~\cite{hu2021lora} structure for fine-tuning. For each layer's LoRA component, we set $r = 16$ and $\alpha = 8$.  We train the model for 250 epochs~\cite{zanella2024low}, using two hyperparameters, $\beta$ and $\lambda$ (refer to Method). For all settings, we fix $\beta = 7$ and $\lambda = 50$. We evaluate each model using 15 query samples per class, randomly selecting 800 episodes, and report the results with a 95\% confidence interval. Reffer to Appendix for more information.

\subsection{Comparison with the state-of-the-arts}
We present the performance of the most representative CDFSL  models~\cite{hu2022adversarial,fu2022wave,fu2023styleadv,zhao2023dual,hu2022pushing,zou2024flatten,zou2025closer,zouattention,yazdanpanah2022visual,xu2024enhancing,xu2024step,zhou2022learning}, domain adaption model~\cite{bai2024prompt} and few-shot learning model~\cite{khattak2023maple,zanella2024low,tang2024amu,huang2024lp++,li2025logits,zhang2021tip}. These models have different backbones~\cite{tschannen2025siglip,bolya2025perception,radford2021learning}. 
Our method, as an extension, can be applied to existing models. Therefore, we selected two existing models—Maple~\cite{khattak2023maple} and CLIP-LoRA~\cite{zanella2024low}—to integrate our method. As in Table~\ref{tab:exp_cdfsl}, applying our method effectively improves their performance, achieving new state-of-the-art results. Also, we further demonstrate our method's effectiveness on the Meta-dataset~\cite{triantafillou2019meta} (Fig.~\ref{fig:meta_dataste}); \textbf{more details and results are provided in the Appendix.}

\begin{figure}[!t]
\centering
\subfloat{
\begin{minipage}[b]{0.95\linewidth}
\begin{adjustbox}{max width=1\linewidth}
\includegraphics[]{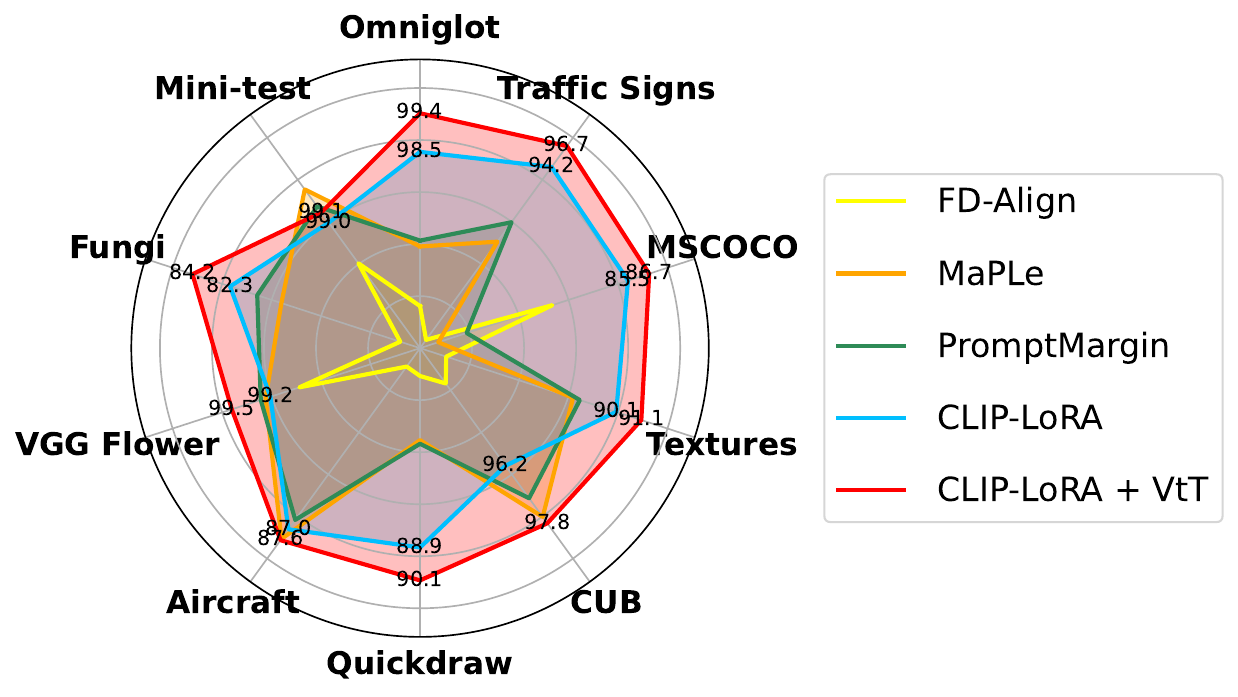}
\end{adjustbox}
\end{minipage}
} 
\caption{The 5-way 5-shot results on 10 Meta-datasts~\cite{triantafillou2019meta}, see Appendix for the 5-way 1-shot and detailed results.}
\label{fig:meta_dataste}
\end{figure}

\begin{figure}[!t]
\centering
\begin{minipage}[b]{0.95\linewidth}
    \centering
    \centering
    \subfloat[a photo of a Industrial Buildings.]{
    \includegraphics[width=0.24\linewidth]{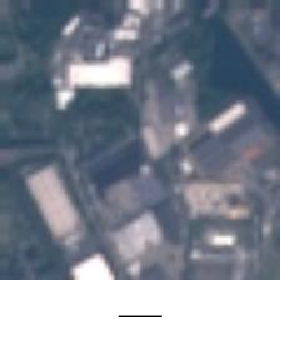} 
    \includegraphics[width=0.24\linewidth]{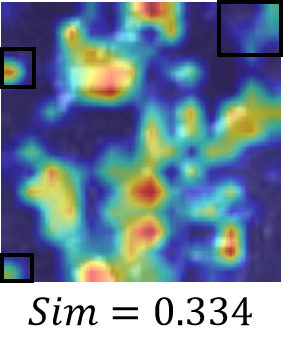} 
    \includegraphics[width=0.24\linewidth]{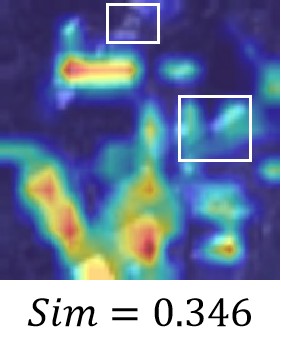} 
    \includegraphics[width=0.24\linewidth]{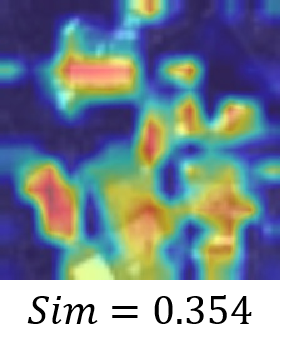} 
    }
\end{minipage}
\\
\begin{minipage}[b]{0.95\linewidth}
    \centering
    \centering
    \subfloat[a photo of a Vascular Lesion.]{
    \includegraphics[width=0.24\linewidth]{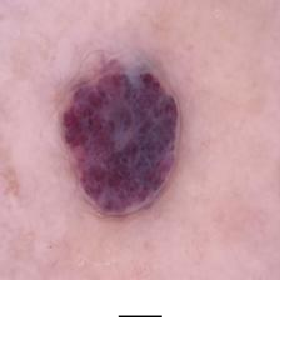} 
    \includegraphics[width=0.24\linewidth]{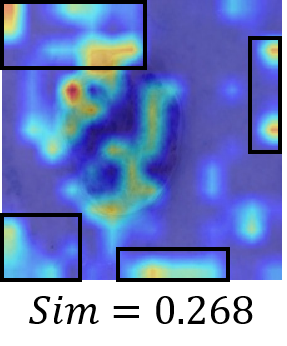} 
    \includegraphics[width=0.24\linewidth]{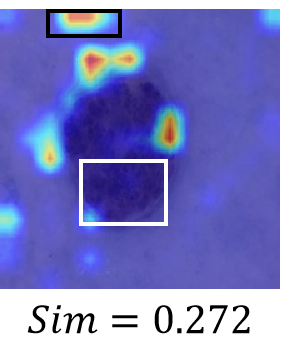} 
    \includegraphics[width=0.24\linewidth]{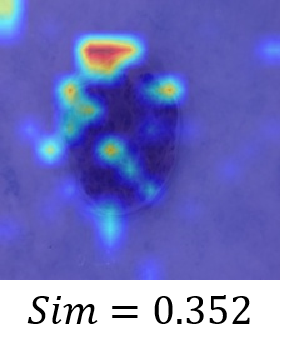}
    }
\end{minipage}
\caption{The attention maps. From left to right: the original image, the baseline result, baseline + remove (see Figure~\ref{fig:2_mthond}) result, and the result of ours. Black boxes highlight areas of incorrect attention, while white boxes highlight missing attention. $Sim$ represents the cosine similarity between the image features and the text features. A higher similarity indicates better alignment. }
\label{fig:heat_map}
\end{figure}

\subsection{Reclaiming the Lost Layer}
As shown in Figure~\ref{fig:intro_lost_ours}, our method reutilizes the lost layer. We illustrate the model's changes before and after leveraging this information with several examples. Figure~\ref{fig:heat_map} displays, from left to right: (1) the original image; (2) using the full text encoder, the model incorrectly focuses on non-semantic parts (black box); (3) removing the lost layer eliminates these incorrect focuses but also removes some effective attention areas (white box); (4) our method eliminates incorrect focuses while preserving effective areas, achieving better feature alignment. \textbf{See the Appendix for more analysis and experiments.}

\subsection{Ablation Study}\label{sec:ablation}
\begin{table}[!t]
\caption{Ablation study on 5-way 1-shot task.}
\centering
    \begin{adjustbox}{max width=0.95\linewidth}
    \begin{tabular}{cccccccc}
    \toprule
    TIA & V-T Fusion & DGSO& Crop. & EuroSAT & ISIC & ChestX & Avg\\
    \midrule
    - & - & - & 84.2 & 81.7 & 36.4 & 21.8 & 55.9\\
    \midrule
    $\checkmark$ & - & - & 85.2  &82.9  & 37.5 & 22.0  & 56.9\\
    \midrule
    $\checkmark$ & $\checkmark$ & - & 85.7  &84.1  & 38.0 & 22.5  & 57.6\\
    \midrule
    $\checkmark$ & - & $\checkmark$ & 86.3  &84.0  & 37.6 & 22.3  & 57.6\\
    \midrule    
    $\checkmark$ & $\checkmark$ & $\checkmark$ & 87.0  &85.0  & 38.2 & 22.7  & 58.2\\
    \bottomrule
\end{tabular}
\label{tab:ablation} 
\end{adjustbox}
\end{table}

\begin{table}[!t]
\caption{Ablation study for V-T Fusion module.}
\centering
    \begin{adjustbox}{max width=0.95\linewidth}
    \begin{tabular}{p{0.25\linewidth}cccccc}
    \toprule
    Method & Crop. & EuroSAT & ISIC & ChestX & Avg\\
    \midrule
    \raggedright \textbf{V-T Fusion} & 87.0 & 85.0 & 38.2 & 22.7 & 58.2\\
    \raggedright \mbox{V-T Fusion(R.)} & 86.2 & 84.5 & 38.1 & 22.2 & 57.7\\
    \raggedright \mbox{V-T Fusion(2D)} & 86.9 & 84.7 & 38.4 & 22.7 & 58.2\\
    \midrule
    \raggedright V Fusion & 86.0  &83.4  & 37.3 & 21.8  & 57.1\\
    \raggedright T Fusion & 86.2  &83.9  & 37.9 & 22.1  & 57.5\\
    \midrule
    \raggedright V-T Mean & 85.9  &83.5  & 36.9 & 21.6  & 57.0\\
    \raggedright T Mean & 85.7  &83.5  & 37.5 & 21.7  & 57.1\\
    \raggedright V Mean & 85.8  &83.6  & 37.3 & 21.5  & 57.0\\
    \bottomrule
    \end{tabular}
    \end{adjustbox} 
\label{tab:scan_method} 
\end{table}

\begin{table}[!t]
\caption{Ablation study for TIA module.}
\centering
    \begin{adjustbox}{max width=0.95\linewidth}
    \begin{tabular}{p{0.25\linewidth}cccccc}
    \toprule
    Method & Crop. & Euro. & ISIC & ChestX & Avg\\
    \midrule
    \raggedright OURS & 87.0 & 85.0 & 38.2 & 22.7 & 58.2\\
    \midrule
    \raggedright (a) use $A_{i}^{\prime}$  & 84.7 & 81.6 & 36.5 & 21.5 & 56.1\\
    \raggedright (b) use $A_{i}^{\prime} + t_i$ & 86.2 & 84.1 & 38.1 & 22.3 & 57.7\\
    \raggedright (c) {\footnotesize Append in $r_{i}^{\prime}$} & 86.5 & 83.2 & 37.6 & 22.4 & 57.4\\
    \bottomrule
    \end{tabular}
    \end{adjustbox}   
\label{tab:tia_ablation} 
\end{table}

\begin{table}[t]
\caption{Ablation study on the Dynamic Loss Combining design in DGSO module.}
\centering
    \begin{adjustbox}{max width=0.95\linewidth}
    \begin{tabular}{p{0.25\linewidth}cccccc}
    \toprule
    Method & Crop. & EuroSAT & ISIC & ChestX & Avg\\
    \midrule
    \raggedright VtT w/ DLC & 87.0 & 85.0 & 38.2 & 22.7 & 58.2\\
    \raggedright VtT w/o DLC   & 85.2 & 83.4 & 37.7 & 22.5 & 57.2 \\
    \bottomrule
    \end{tabular}
    \end{adjustbox} 
\label{tab:slid_window} 
\end{table}

\subsubsection{Module Ablation}
Our method consists of three modules: Visual-Text Layer-Level Fusion (V-T Fusion), Text Encoder Information Absorption (TIA) and Dynamic Gradient Supervised Optimization (DGSO). We select CLIP-LoRA-Vision as the baseline and progressively add our modules. The experiments are carried out in the 5-way 1 shot setting on four CDFSL datasets, and the results are given in Table~\ref{tab:ablation}.  As shown, all modules effectively improve the model's performance across the four datasets. When all modules are used together, the model achieves the best performance.

\subsubsection{V-T Fusion Module Ablation}
We compare the V-T Fusion module with alternative feature fusion methods, as shown in Table~\ref{tab:scan_method}. These include: (1) different scanning direction: shallow-to-deep (second row) and bidirectional (third row); (2) single-modal fusion: single visual feature fusion (fourth row) and single text feature fusion (fifth row); (3) average-based fusion: bimodal feature averaging (sixth row), single text feature averaging (seventh row), and single visual feature averaging (eighth row). It can be seen that our V-T Fusion method (first row) consistently outperforms the others. \textbf{See the Appendix for detailed information and the ablation study for the SSM network (Equation~\ref{eq:ssm}): Multi-Head Attention (57.2), RNN (57.2), LSTM (57.4), and SSM (58.2).}


\subsubsection{TIA Module Ablation}
During testing, we remove the VtT network to ensure efficient inference. We also evaluated the performance of using the VtT network for inference by replacing or weighting the class text feature \( t_i \) with \( (a) A_{i}^{\prime} \) (Equation~\ref{eq:gen_a}) or \( (b) A_{i}^{\prime} + t_i \). As in Table~\ref{tab:tia_ablation}, both strategies result in reduced performance.

Additionally, we tested generating \( r^{\prime} \) (Equation~\ref{eq:gen_r}) using the “Append” strategy by concatenating \( A_{i}^{\prime} \) and \( \mathrm{[CLASS]}_5 \), then mapping through a linear network. The results in Table~\ref{tab:tia_ablation} indicate that the Append strategy is less effective than the original direct replacement strategy.
\subsubsection{DGSO Module Ablation}
We analyze the effectiveness of the Dynamic Loss Combining strategy (Equations~\ref{eq:que}-\ref{eq:comb}) within the DGSO module, with results presented in Table~\ref{tab:slid_window}. It is evident that the Dynamic Loss Combining (DLC) strategy significantly enhances the model's performance.

\begin{figure}[!t]
\centering
\subfloat{
\begin{minipage}[b]{0.46\linewidth}
\begin{adjustbox}{max width=1\linewidth}
\includegraphics[]{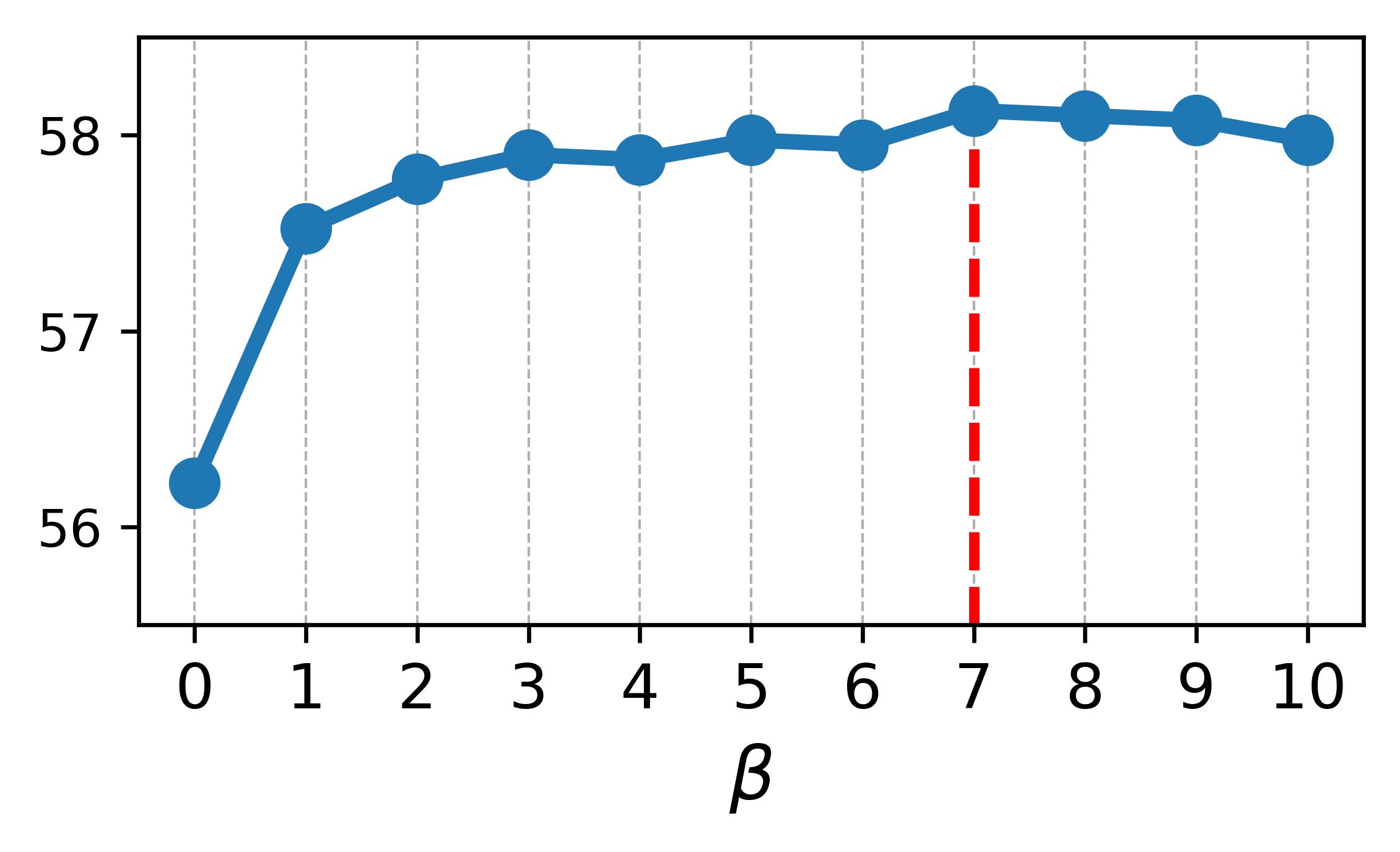}
\end{adjustbox}
\end{minipage}
} 
\subfloat{
\begin{minipage}[b]{0.46\linewidth}
\begin{adjustbox}{max width=1\linewidth}
\includegraphics[]{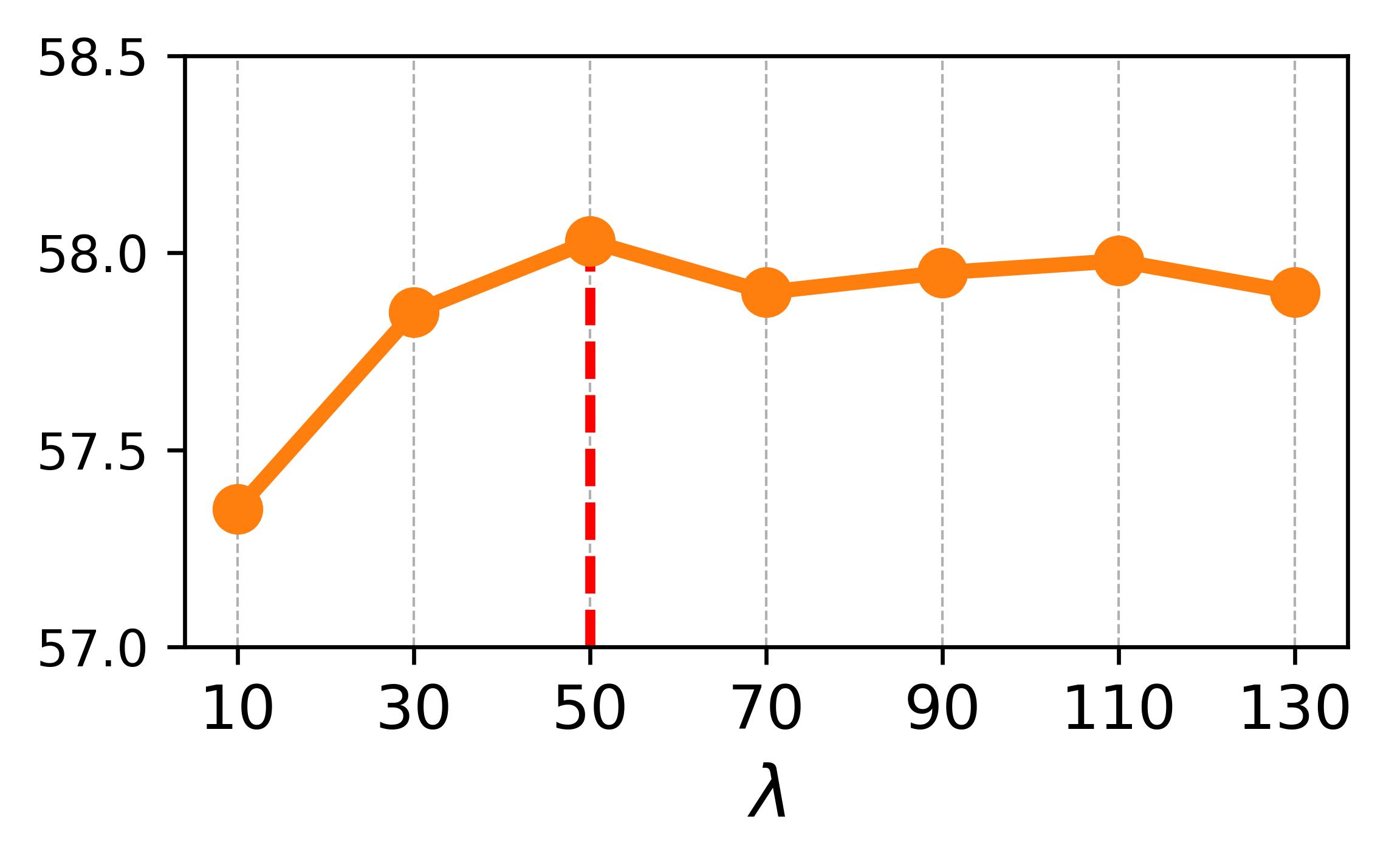}
\end{adjustbox}
\end{minipage}
} 
\caption{Hyperparameter experiments for $\beta$ and $\lambda$. The vertical axis represents the average performance on 5-way 1-shot tasks across four datasets.}
\label{fig:hyper}
\end{figure}

\begin{table}[!t]
\caption{Overhead: CLIP-LoRA-Vision + OURS vs Maple.}
\centering
    \begin{adjustbox}{max width=0.90\linewidth}
    \begin{tabular}{p{0.25\linewidth}cccc}
    \toprule
    Method & Parameter (M) & FLOPs (G) & Acc \\
    \midrule
    \raggedright Maple~\cite{khattak2023maple} & 3.1 & 205 & 53.1 \\
    \raggedright Ours & 3.9 & 148.5 & 58.2 \\
    \bottomrule
    \end{tabular}
    \end{adjustbox} 
\label{tab:over_head} 
\end{table}

\subsection{Hyperparameter Study and Overhead.}
Our model includes two hyperparameters, $\beta$ and $\lambda$ (see the DGSO section in the Method part). We conduct hyperparameter ablation experiments, as shown in Figure~\ref{fig:hyper}. It can be seen that the best performance is achieved when $\beta = 7$ and $\lambda = 50$. Also, our method maintains a low computational overhead, as demonstrated in Table~\ref{tab:over_head}.



%% file: Tex/07_con.tex
\section{Conclusion}
In this paper, we observe the Lost Layer in CLIP under cross-domain scenarios and reveal that the information within the Lost Layer is actually beneficial for SF-CDFSL tasks. However, changes in the visual domain lead to the under-utilization of this information. We then propose the VtT model to reutilize this information. Extensive experiments validate our rationale and effectiveness.

%% file: Tex/08_sup.tex
\appendix

\section{Text Encoder is Better for Cross-Domain Task}
In the main text, we described the discovery of the lost layer and proposed the VtT model to address this issue. In this section, we provide additional insights into our previous findings and the design of our method. We conducted an in-depth comparison between the text encoder and the vision encoder of CLIP, examining their differences and their impact in cross-domain scenarios.

\subsection{Different Thinking Patterns}
We examined the attention distribution corresponding to the CLS/EOS token in each layer of the CLIP's vision and text encoder , as illustrated in Figure~\ref{fig:bef_1a} and~\ref{fig:bef_1c}. We have two discoveries:
(1) In the vision encoder, shallow layers focus on specific semantic parts, while deeper layers shift to focusing on a small number of background non-semantic tokens.
(2) In the text encoder, shallow layers concentrate on contextual content, whereas deeper layers shift to focusing on semantically rich category tokens.

We also quantitatively demonstrated this phenomenon. We measured the attention weight ratio of category tokens to all text tokens in each layer of the text encoder, as depicted by the red line in Figure~\ref{fig:bef_1d}. The attention weight of the EOS token to category tokens in shallow layers is nearly zero, indicating that these blocks predominantly focus on context rather than task-relevant semantic information. As the layers deepen, this attention ratio increases, with over 60\% of attention in the final layers directed towards aggregating category-related token information. Similarly, in the vision branch, we tracked the attention weight of the top 10 tokens with the highest attention in the final layer across all layers, shown by the red line in Figure~\ref{fig:bef_1b}. As the layers deepen, these top 5\% tokens capture approximately 60\% of the attention score.

\subsection{How This Difference Influences Cross-Domain Tasks}
\begin{figure*}[h]
\centering
\subfloat[]{
	\label{fig:bef_1a}\includegraphics[width=0.66\linewidth]{./Figs/bef_a.png}}
\subfloat[]{
	\label{fig:bef_1b}\includegraphics[width=0.32\linewidth]{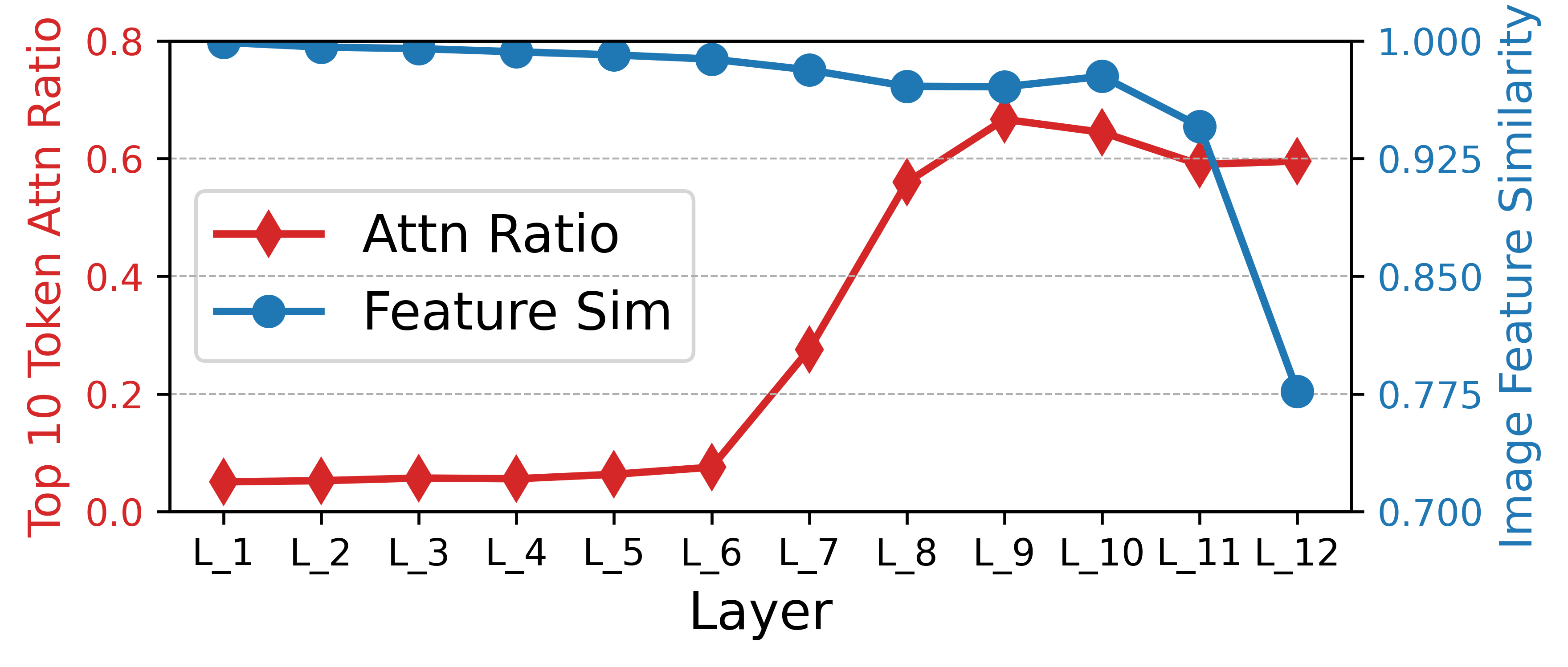}}
\\
\subfloat[]{
	\label{fig:bef_1c}\includegraphics[width=0.66\linewidth]{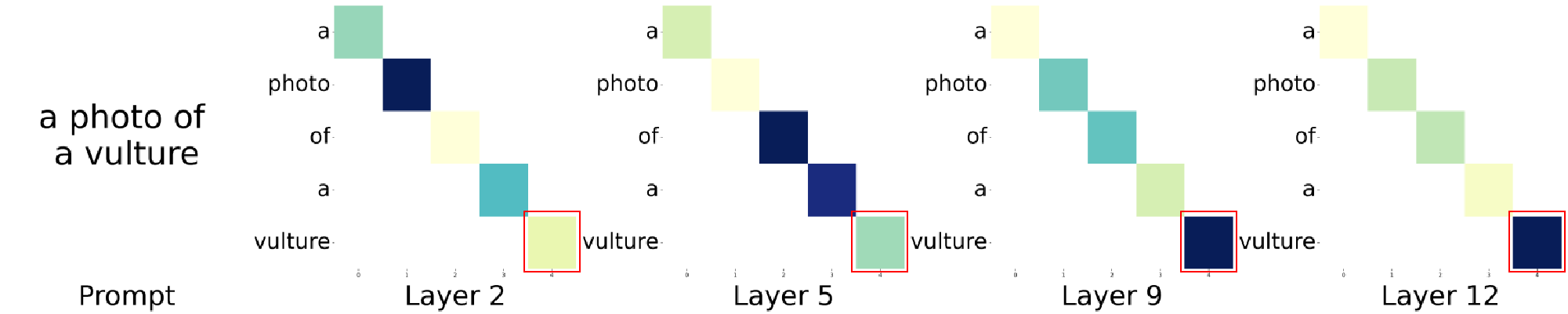}} 
\subfloat[]{
	\label{fig:bef_1d}\includegraphics[width=0.32\linewidth]{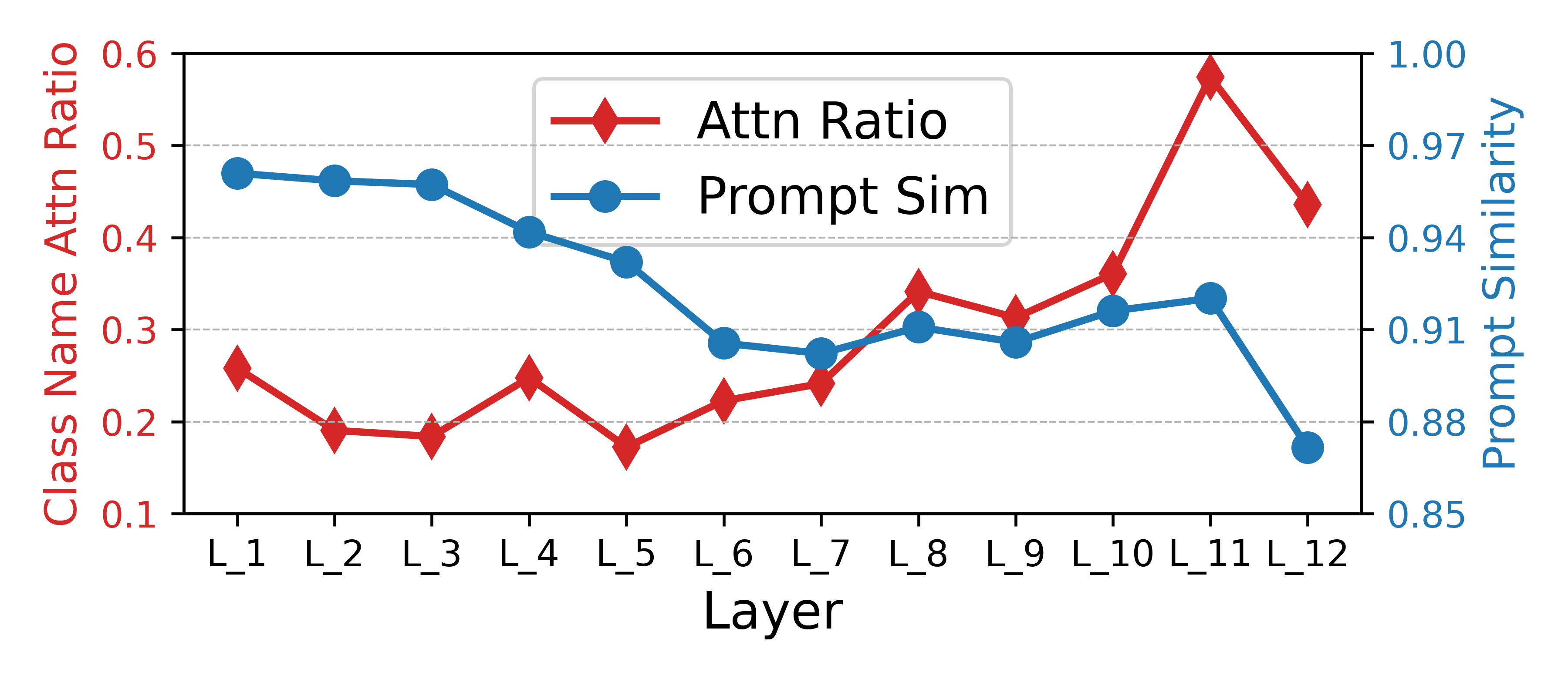}} 
\\
    \subfloat[]{
     \label{fig:text_better_a}\includegraphics[width=0.48\linewidth]{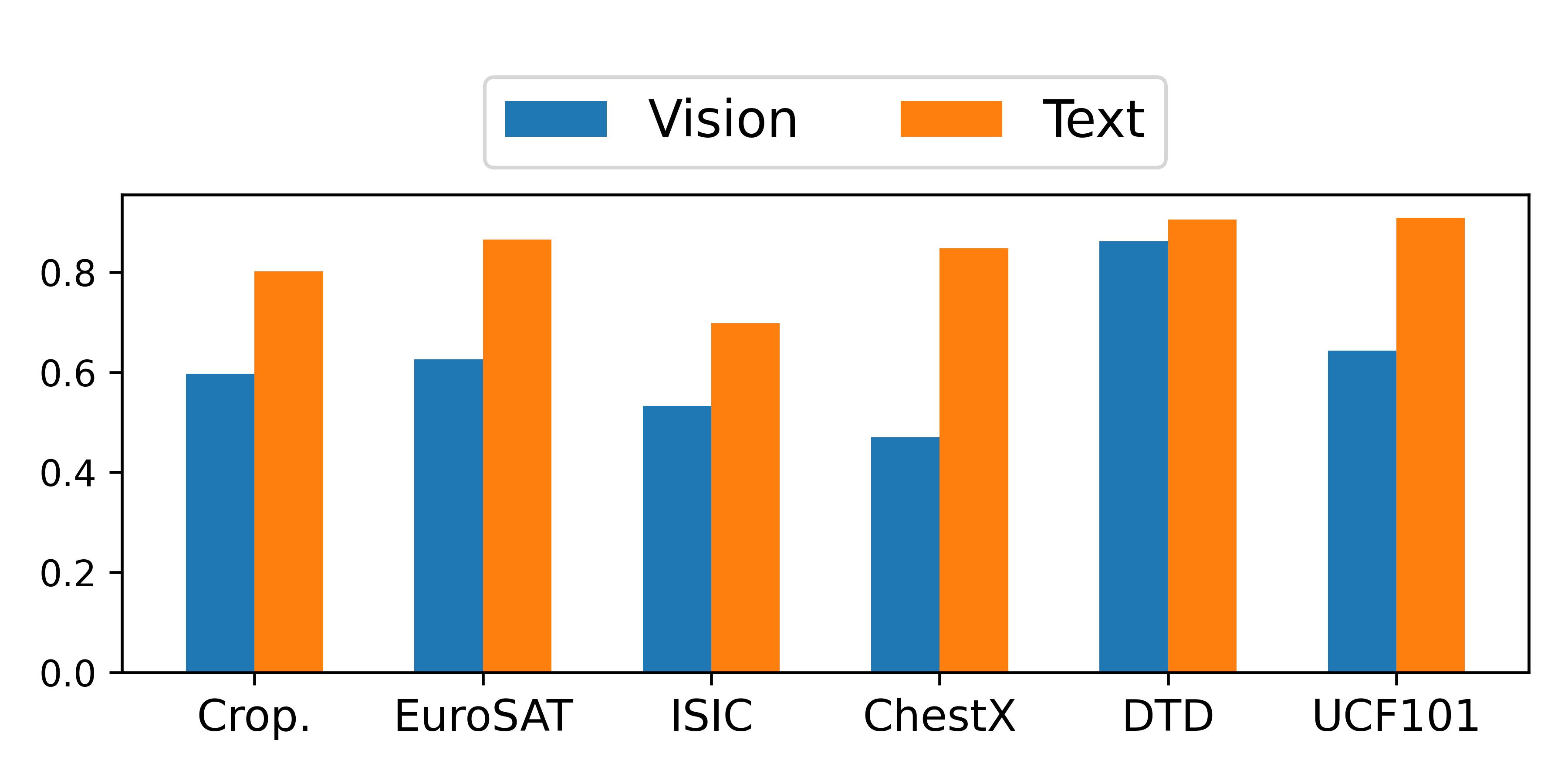}} 
    \subfloat[]{
	\label{fig:text_better_b}\includegraphics[width=0.48\linewidth]{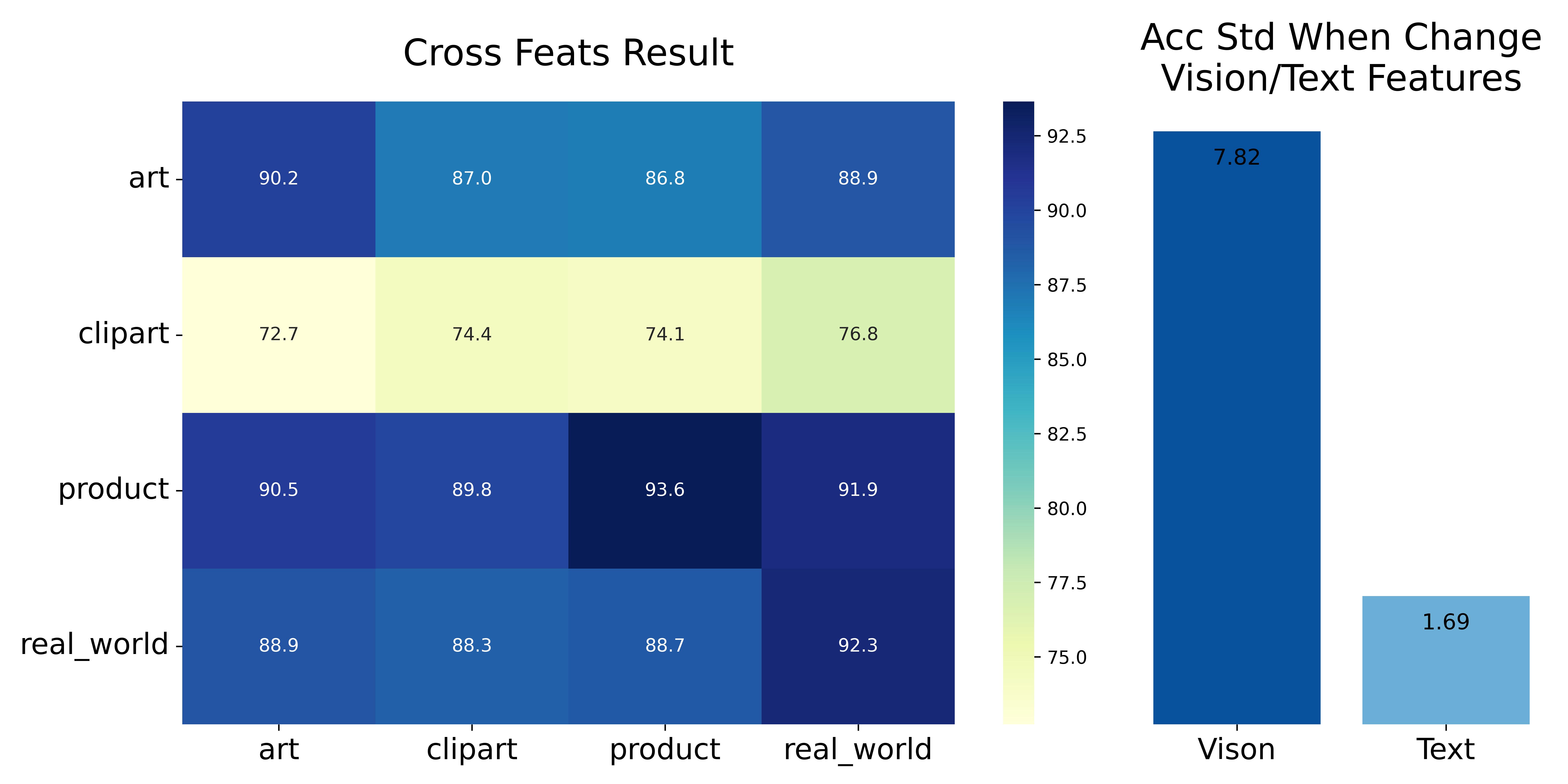}} 
\caption{(a) The attention distribution across different layers of the vision encoder. (b) The red line quantitatively shows the attention weight percentage of the top 10 tokens with the highest attention weights in each layer, while the blue line shows the similarity of features extracted by different layers of the vision encoder for visual samples from different domains but of the same category. A higher similarity indicates more domain-independent features. (c) Visualization of the attention weight distribution across different layers of the text encoder. (d) The red line quantitatively shows the attention weight percentage of the category name token in each layer of the text encoder, while the blue line shows the similarity of features extracted by different layers of the text encoder for textual descriptions with different domain information but containing the same category name. A higher similarity indicates more domain-independent features. (e) Domain distance between visual features and text features extracted by the CLIP model when transferred from the ImageNet dataset (source domain) to downstream cross-domain datasets. (f) Cross-domain transfer experiments of visual and textual features from different domains.}
\end{figure*}

An intuitive hypothesis is that a text encoder model, which focuses more on semantic information in the final layers, should extract more domain-independent features. For instance, the phrases “a photo of a dog” and “a photo of a cartoon dog” should emphasize the semantic information “dog” thereby mitigating the impact of domain-specific information like “cartoon”. To validate this hypothesis, we measured the similarity of text and visual features from different layers of the encoder for the same category but from different domains on ImageNet-R. We use different templates containing domain information to simulate cross-domain conditions in the text branch, such as “a cartoon photo of a dog” and “a photo of a painting-style dog”. The results are shown by the blue line in Figure~\ref{fig:bef_1b} and~\ref{fig:bef_1d}.

In Figure~\ref{fig:bef_1d}, the similarity of text features containing different domain information declines in the initial layers but increases in the final layers, supporting our hypothesis. The text encoder in the final layers (believed to emphasize classification tasks) stresses category-related semantic information. Consequently, the similarity of text descriptions with the same semantics but different domain information increases in these layers. The final extracted text characteristics, which are \textbf{ dominated by semantic information, are less sensitive to domain differences}. The similarity between text features of the same category but different domains remains above 0.9, indicating that the text encoder extracts domain-independent features.

Conversely, the vision encoder reduces its focus on semantic parts in the final layers and shifts attention to a small number of background tokens (Figure~\ref{fig:bef_1b}, red line). The similarity between features of the same category but different domains is nearly 1 in the initial layers but significantly decreases in the final layers. This indicates that the vision encoder's “domain-sensitive” approach in the final layers amplifies the impact of domain differences, resulting in a similarity of around 0.77, much lower than the text encoder's 0.9.

\subsection{Experimental Proof}
We also conducted specific experiments to demonstrate that, compared to the vision encoder, the text encoder in CLIP contains more domain-independent information, making it easier to transfer to cross-domain tasks.

We utilized the CKA~\cite{kornblith2019similarity} (Centered Kernel Alignment) similarity metric to measure domain distance between visual features and text features extracted by the CLIP model when transferred from the ImageNet dataset (source domain) to downstream cross-domain datasets. A higher CKA similarity indicates a smaller domain distance, suggesting that the encoder contains less domain-specific information. As shown in Figure~\ref{fig:text_better_a}, the CKA metric for text features is consistently higher than that for visual features across all downstream datasets, indicating that the text encoder extracts more domain-independent text features.

To further demonstrate the robustness of the text encoder against domain information, we conducted cross-domain experiments with visual and text features. We selected the Office-Home dataset~\cite{venkateswara2017deep}, which contains category names and visual images of the same categories from four different domains. Using LoRA, we fine-tuned the vision and text branches separately on the datasets from the four domains, recorded the resulting visual and text features, and performed cross-combination CLIP classification tasks. The results, shown in Figure~\ref{fig:text_better_b}, indicate that when we fix the visual features (vertical axis) and vary the text features (horizontal axis), the performance change is minimal. This implies that the text features extracted from different domains are relatively consistent and have good transferability. Conversely, when we fix the text features and vary the visual features (viewed from top to bottom), the performance varies significantly, indicating that the visual features learned from different domains are vastly different and less transferable. The right part of Figure~\ref{fig:text_better_b} illustrates the standard deviation in classification accuracy as visual and text features change.

\subsection{Conclusion and Discussion}
Through our analysis, we found that compared to the vision encoder, the text encoder in CLIP: (1) operates in a semantically-driven manner, making it less sensitive to domain information, and (2) its information transfers better to cross-domain tasks. These findings further validate our proposed idea of “teaching the vision encoder to think like the text encoder”, emphasizing its significance and necessity in cross-domain scenarios.

\begin{table}[t]
\centering
\caption{In various backbone versions of the CLIP model, masking a specific layer of the text encoder can lead to significant performance improvements in zero-shot SF-CDFSL tasks.}
\label{tab:intro_zero} 
\begin{adjustbox}{max width=\linewidth}
\begin{tabular}{*{9}{c}}
    \toprule 
     \multirow{2}*{Backbone}  
       & \multicolumn{2}{c}{CropDisease}  & \multicolumn{2}{c}{EuroSAT}  & \multicolumn{2}{c}{ISIC}   & \multicolumn{2}{c}{ChestX}                                 \\
     \cmidrule(lr){2-3}\cmidrule(lr){4-5}\cmidrule(lr){6-7}\cmidrule(lr){8-9}
     & Full & Masked & Full & Masked & Full & Masked & Full & Masked  \\
    \midrule 
    VIT-RN50 & 34.7 & 37.8\scriptsize\textcolor{red}{+3.1} &38.0 &39.4\scriptsize\textcolor{red}{+1.4} &24.7 & 28.1\scriptsize\textcolor{red}{+3.4} & 19.9 & 20.1\scriptsize\textcolor{red}{+0.2} \\
    VIT-RN101 & 34.1 & 35.6\scriptsize\textcolor{red}{+1.5} &38.1 &40.3\scriptsize\textcolor{red}{+2.2} &27.6 & 27.7\scriptsize\textcolor{red}{+0.1} & 19.6 & 20.0\scriptsize\textcolor{red}{+0.4} \\
    VIT-B\/16 & 39.8 & 44.0\scriptsize\textcolor{red}{+4.2} &54.8 &63.2\scriptsize\textcolor{red}{+8.4} &27.3 & 29.2\scriptsize\textcolor{red}{+1.9} & 21.8 & 21.9\scriptsize\textcolor{red}{+0.1} \\
    VIT-L\/14 & 52.2 & 55.2\scriptsize\textcolor{red}{+3.0} &69.7 &71.2\scriptsize\textcolor{red}{+1.5} &27 & 29.8\scriptsize\textcolor{red}{+2.8} & 21.7 & 22.5\scriptsize\textcolor{red}{+0.8} \\
    \bottomrule 
\end{tabular}
\end{adjustbox}
\end{table}

\begin{table}[t]
\centering
\caption{When employing different PEFT methods for 5-way 1-shot SF-CDFSL tasks, training after removing a specific layer of the text encoder can achieve better performance compared to using the complete text encoder.}
\label{tab:intro_peft} 
\begin{adjustbox}{max width=\linewidth}
\begin{tabular}{*{9}{c}}
    \toprule 
     \multirow{2}*{Method}  
       & \multicolumn{2}{c}{CropDisease}  & \multicolumn{2}{c}{EuroSAT}  & \multicolumn{2}{c}{ISIC}   & \multicolumn{2}{c}{ChestX}                                 \\
     \cmidrule(lr){2-3}\cmidrule(lr){4-5}\cmidrule(lr){6-7}\cmidrule(lr){8-9}
     & Full & Masked & Full & Masked & Full & Masked & Full & Masked  \\
    \midrule 
    Lora-Vison & 82.5 & 85.0\scriptsize\textcolor{red}{+2.5} &81.8 &82.4\scriptsize\textcolor{red}{+0.6} &35.4 & 36.3\scriptsize\textcolor{red}{+0.9} & 21.5 & 21.8\scriptsize\textcolor{red}{+0.3} \\
    Lora-Text & 83.1 & 83.3\scriptsize\textcolor{red}{+0.2} &74.1 &74.6\scriptsize\textcolor{red}{+0.5} &34.4 & 34.5\scriptsize\textcolor{red}{+0.1} & 21.4 & 22.2\scriptsize\textcolor{red}{+0.8} \\
    Lora-Both & 84.3 & 84.6\scriptsize\textcolor{red}{+0.3} &81.4 &82.9\scriptsize\textcolor{red}{+1.5} &33.6 & 34.3\scriptsize\textcolor{red}{+0.7} & 22.4 & 22.7\scriptsize\textcolor{red}{+0.3} \\
    Maple & 81.8 & 82.6\scriptsize\textcolor{red}{+0.8} &76.5 &77.2\scriptsize\textcolor{red}{+0.7} &33.6 & 33.8\scriptsize\textcolor{red}{+0.2} & 21.3 & 22.2\scriptsize\textcolor{red}{+0.9} \\
    \bottomrule 
\end{tabular}
\end{adjustbox}
\end{table}

\section{Widespread Occurrence of the Lost Layer}
We find that the phenomenon of shortcut layers is prevalent in CLIP of different backbones, as shown in Table~\ref{tab:intro_zero}. Moreover, this phenomenon persists even after fine-tuning with a few samples, as seen in Table~\ref{tab:intro_peft}. It is evident that removing a certain layer from the text encoder (Masked) consistently improves classification performance across four cross-domain few-shot datasets.
\section{Detailed V-T Fusion Module Ablation}
Additionally, we demonstrate the performance of various methods for integrating the outputs of the text encoder and visual encoder to highlight the rationality and effectiveness of our V-T Fusion module. First, we compare the performance of different scanning strategies used in the V-T Fusion module to generate ${L_i}$, as shown in the first three rows of Table~\ref{tab:scan_method}. Our approach, V-T Fusion, uses a deep-to-shallow scanning strategy. In contrast, V-T Fusion(R.) represents the reverse, using a shallow-to-deep scanning strategy. V-T Fusion(2D) involves scanning in both directions, with both sequences fed into the subsequent SSM network, and their outputs combined. As shown, V-T Fusion achieves the best performance. Although V-T Fusion(2D) yields similar results, considering computational complexity, we opt for the single-scan V-T Fusion method.

We also compare other methods of integrating visual and text features from each layer, with results shown in rows 4 to 8 of Table~\ref{tab:scan_method}. T Fusion and V Fusion denote learning with only the text output and visual input, respectively, into the SSM network. V-T Mean, T Mean, and V Mean denote averaging the outputs of all layers from both branches, only the text branch, and only the visual branch, respectively. It is evident that none of these methods outperform our proposed V-T Fusion approach.

\begin{table}[t]
\caption{Further ablation study for V-T Fusion module on 5-way 1-shot task.}
\label{tab:scan_method} 
\centering
    \begin{adjustbox}{max width=0.95\linewidth}
    \begin{tabular}{p{0.25\linewidth}cccccc}
    \toprule
    Method & Crop. & EuroSAT & ISIC & ChestX & Avg\\
    \midrule
    \raggedright \textbf{V-T Fusion} & 87.0 & 85.0 & 38.2 & 22.7 & 58.2\\
    \raggedright \mbox{V-T Fusion(R.)} & 86.2 & 84.5 & 38.1 & 22.2 & 57.7\\
    \raggedright \mbox{V-T Fusion(2D)} & 86.9 & 84.7 & 38.4 & 22.7 & 58.2\\
    \midrule
    \raggedright V Fusion & 86.0  &83.4  & 37.3 & 21.8  & 57.1\\
    \raggedright T Fusion & 86.2  &83.9  & 37.9 & 22.1  & 57.5\\
    \midrule
    \raggedright V-T Mean & 85.9  &83.5  & 36.9 & 21.6  & 57.0\\
    \raggedright T Mean & 85.7  &83.5  & 37.5 & 21.7  & 57.1\\
    \raggedright V Mean & 85.8  &83.6  & 37.3 & 21.5  & 57.0\\
    \bottomrule
    \end{tabular}
    \end{adjustbox} 
\end{table}

\begin{table}[!t]
\caption{Ablation study for SSM network.}
\centering
    \begin{adjustbox}{max width=0.95\linewidth}
    \begin{tabular}{p{0.25\linewidth}cccccc}
    \toprule
    Method & Crop. & Euro. & ISIC & ChestX & Avg\\
    \midrule
    \raggedright SSM (OURS) & 87.0 & 85.0 & 38.2 & 22.7 & 58.2\\
    \raggedright MH Attention  & 85.3 & 84.2 & 37.3 & 22.1 & 57.2\\
    \raggedright RNN & 85.9 & 83.9 & 36.9 & 22.2 & 57.2\\
    \raggedright LSTM & 85.8 & 84.3 & 37.6 & 22.1 & 57.4\\
    \bottomrule
    \end{tabular}
    \end{adjustbox} 
\label{tab:ssm_ablation} 
\end{table}

\begin{figure*}[!t]
\centering
\subfloat[]{
\begin{minipage}[b]{0.48\linewidth}
\label{fig:seman_inf}\includegraphics[width=1\linewidth]{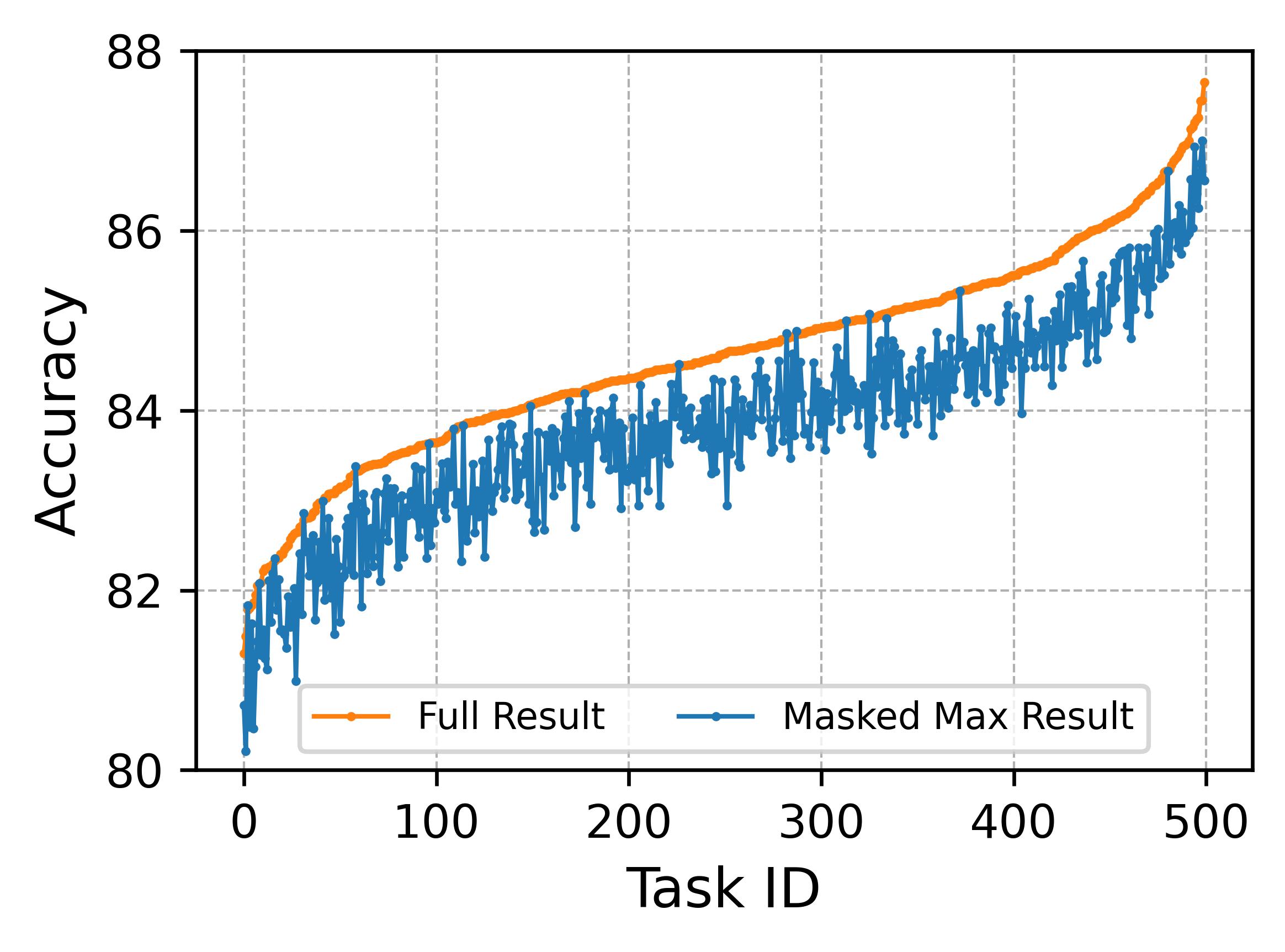}
\end{minipage}}
\medskip
\subfloat[]{
\label{fig:v_t_sim}
\begin{minipage}[b]{0.46\linewidth}
\includegraphics[width=0.49\linewidth]{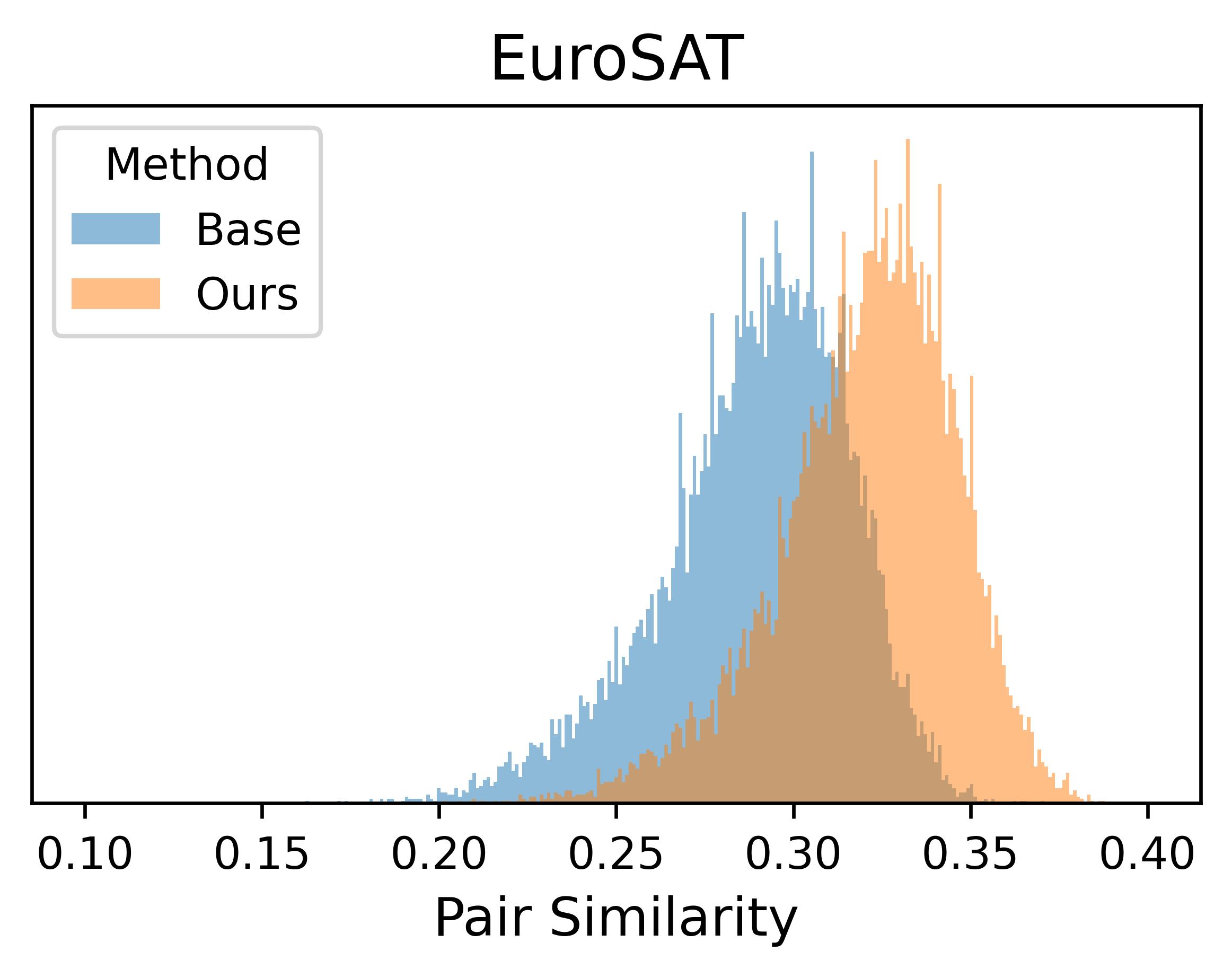}
\includegraphics[width=0.49\linewidth]{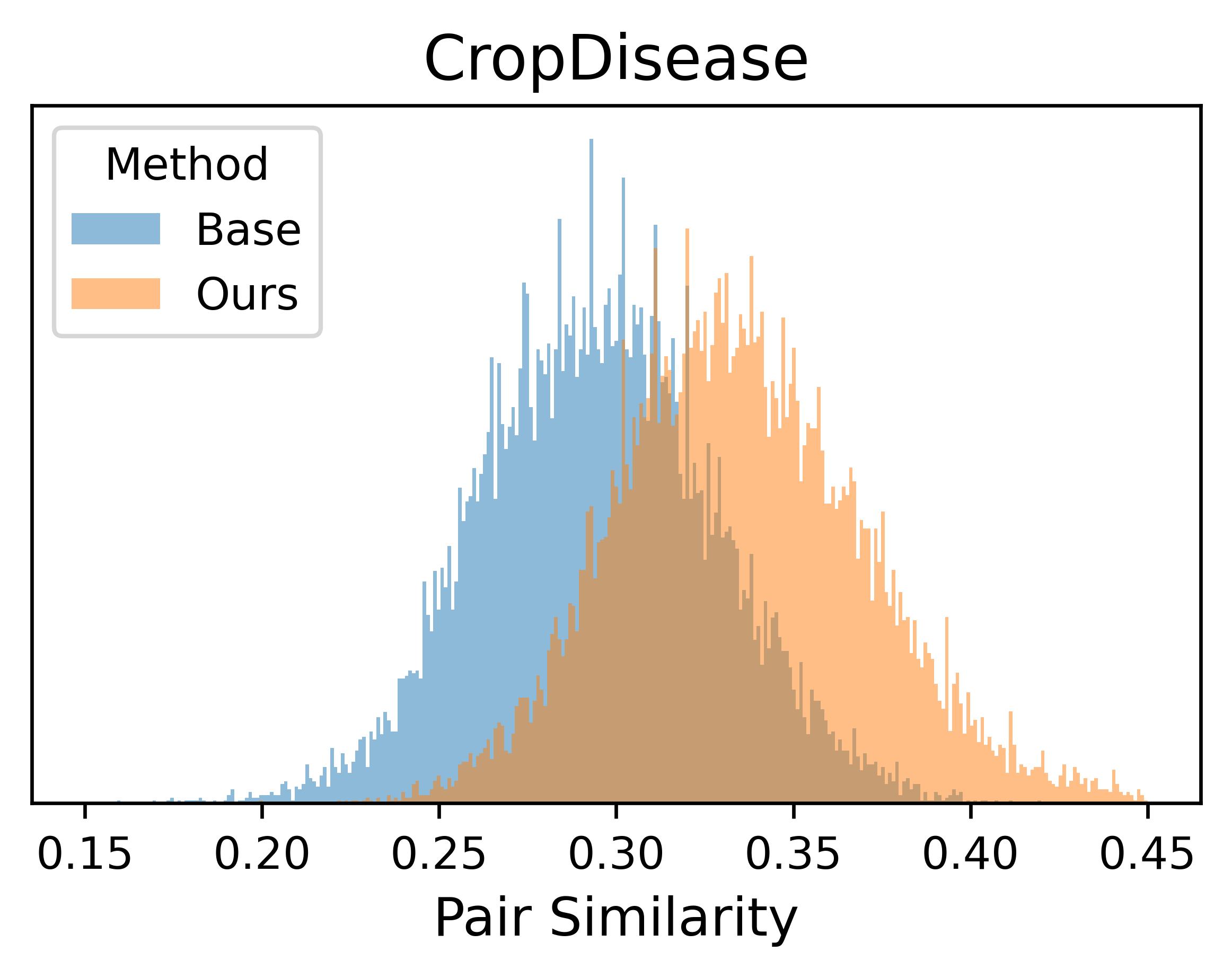}
\\
\includegraphics[width=0.49\linewidth]{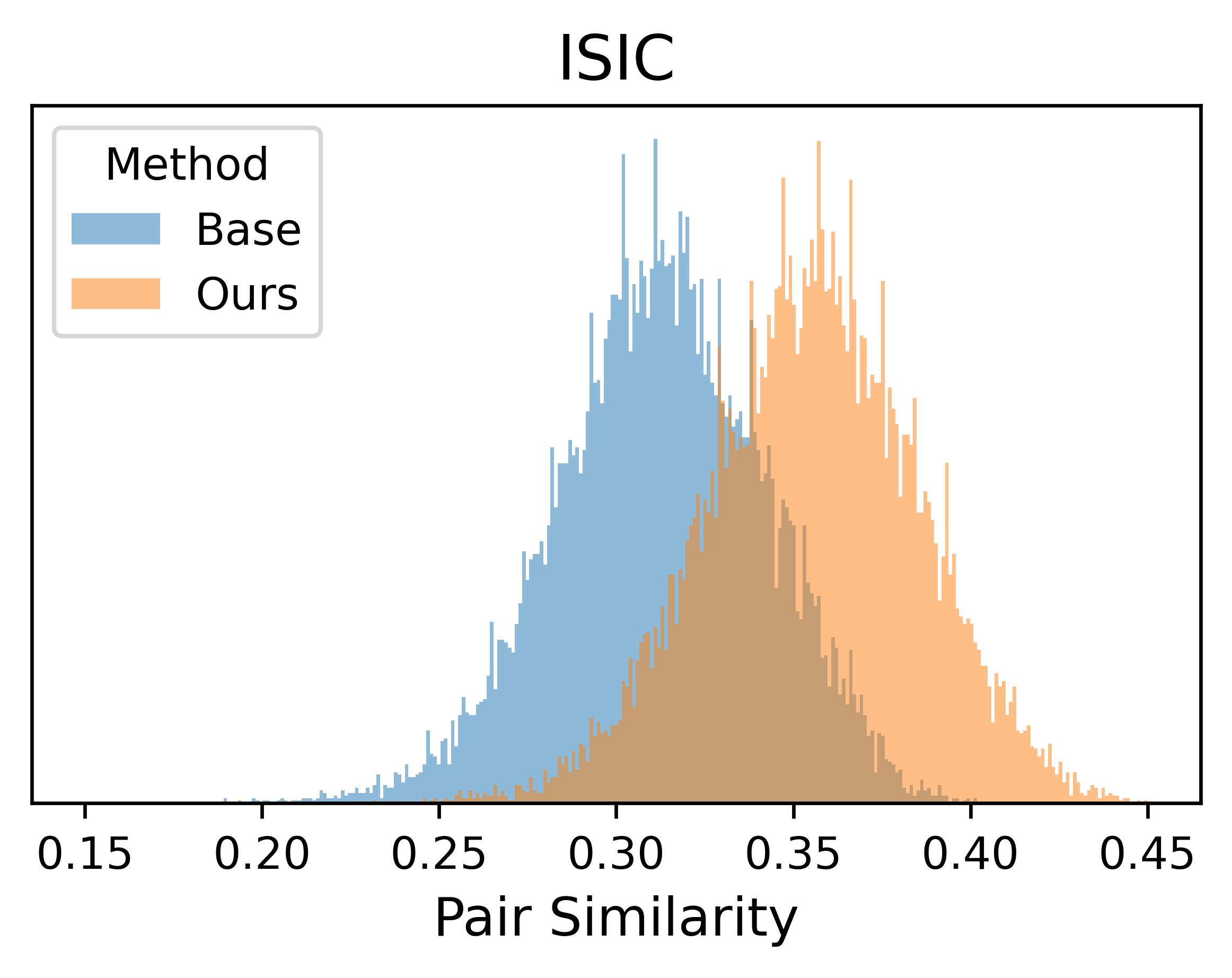}
\includegraphics[width=0.49\linewidth]{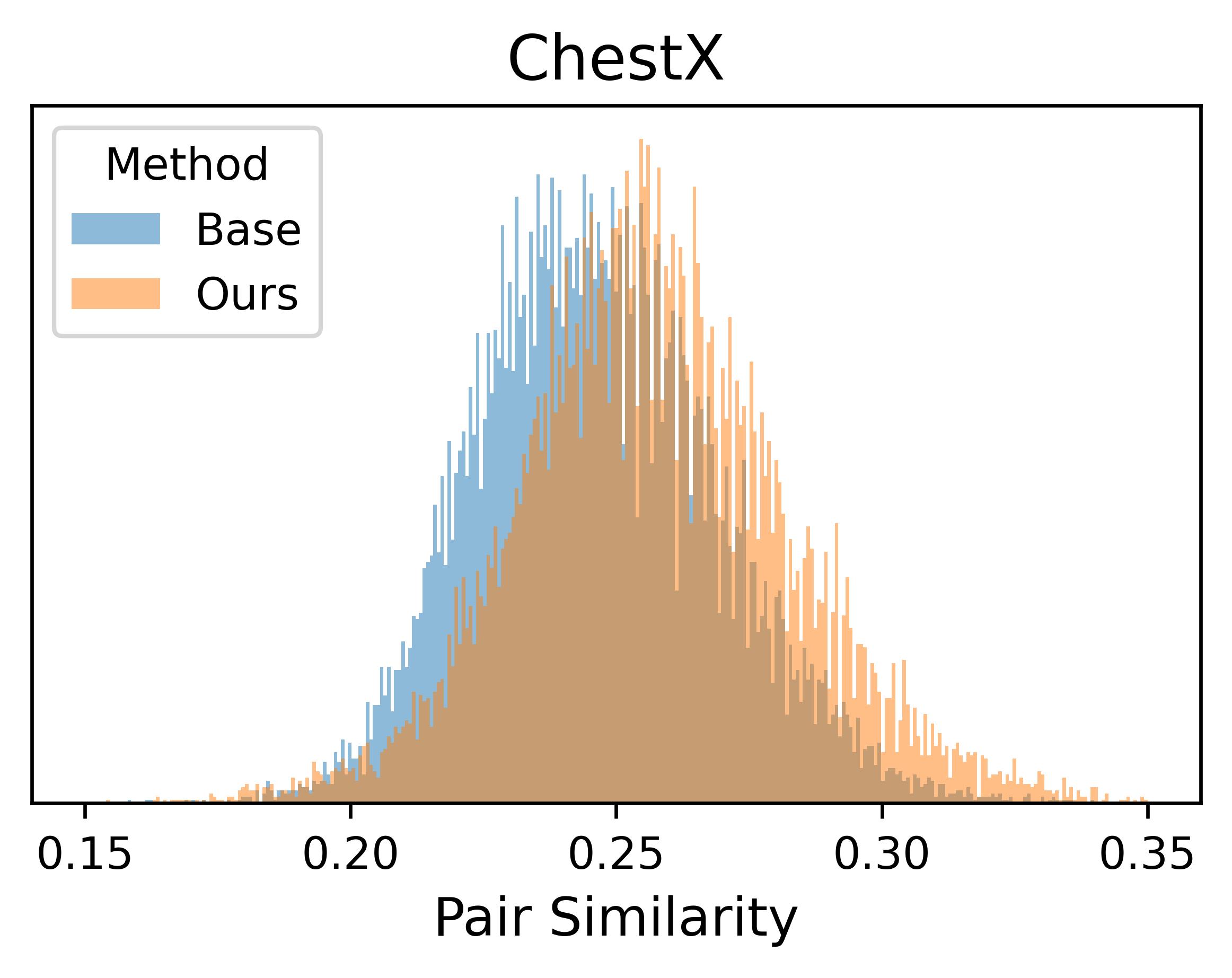}
\end{minipage}   
}
\caption{(a) In the source domain (ImageNet), performance on sampled sub-tasks containing 100 different categories shows that optimal performance is almost always achieved using the complete text encoder, suggesting that the category composition is not the cause of the shortcut layer. (b) Cosine similarity between the visual features and the corresponding textual features extracted by the model before and after applying our method.}
\end{figure*}

\begin{figure}[!t]
\centering
\subfloat[base]{
	\label{fig:tsne_a}\includegraphics[width=0.47\linewidth]{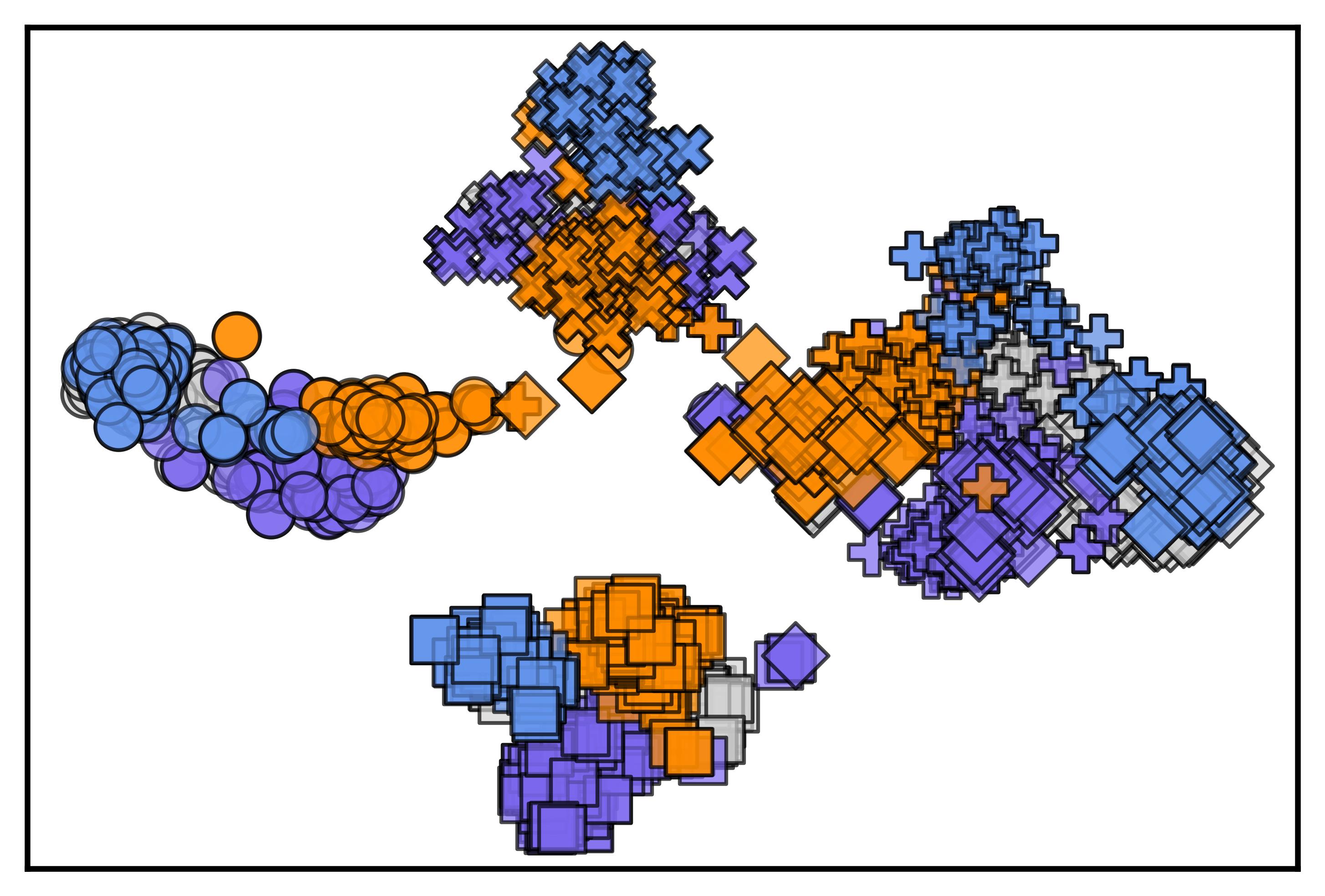}}
\subfloat[ours]{
	\label{fig:tsne_b}\includegraphics[width=0.47\linewidth]{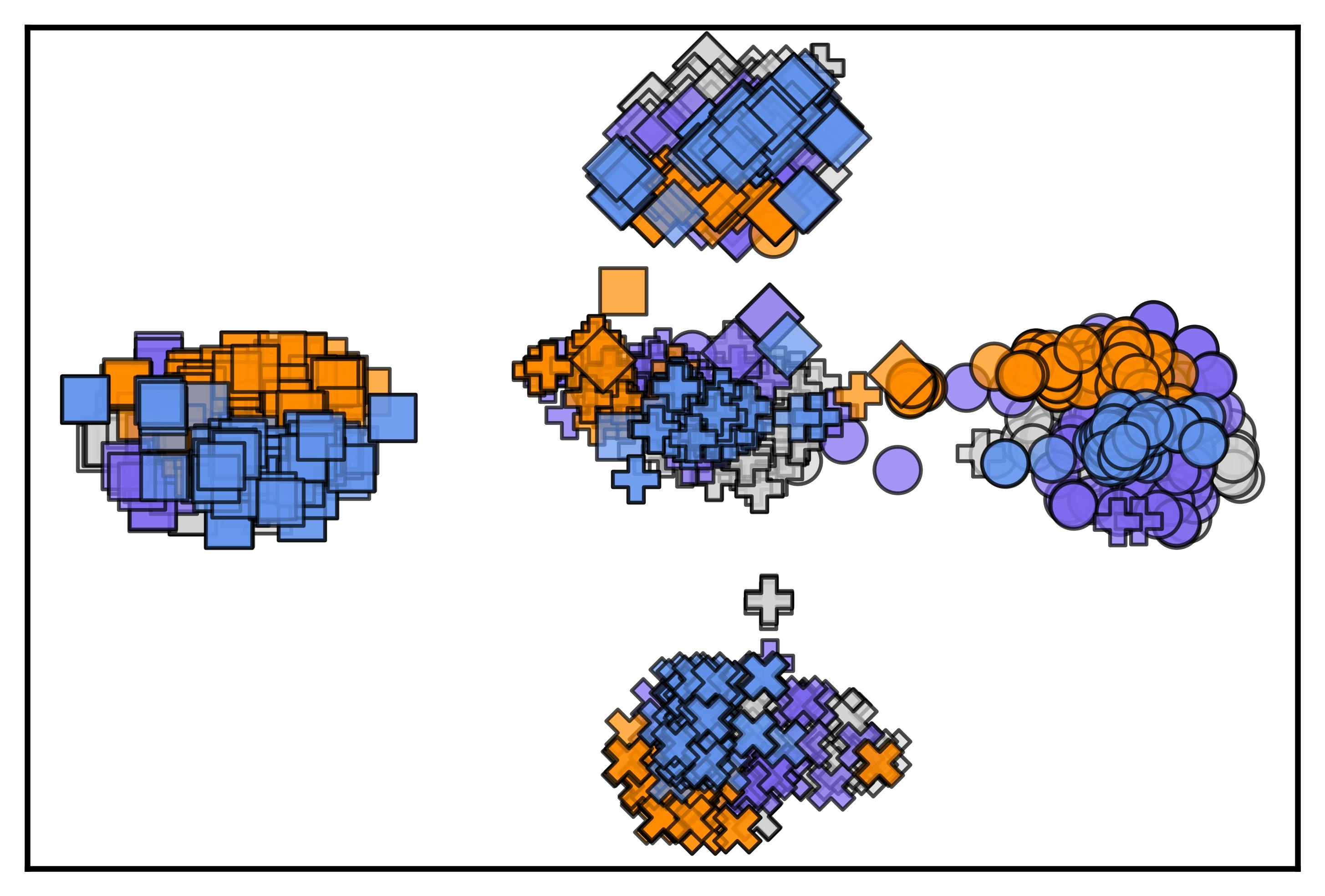}}
\caption{t-SNE~\cite{maaten2008visualizing} visualization results for the ImageNet-R~\cite{hendrycks2021many} (5 classes: glodfish, hammerhead shark, tiger shark, stingray and hen). Different colors denote samples from different domain, and different shapes indicate samples of different classes. }
\label{fig:tsne}
\end{figure}

\section{Replacing SSM Network}
In the V-T Fusion module, we use the SSM network to serialize and fuse information from the visual-text feature sequences (see Equation~\ref{eq:ssm} in the main text). In this section, we further evaluate the performance of other methods that replace the SSM network. The comparison methods mainly fall into two categories: the first category includes sequential modeling methods such as RNN~\cite{zaremba2014recurrent} and LSTM~\cite{shi2015convolutional}. The second category includes attention mechanism networks like Multi-Head Attention~\cite{vaswani2017attention}. The results are shown in Table~\ref{tab:ssm_ablation}. It can be seen that none of these methods outperforms the SSM network used by us.

\section{Semantic Information Does Not Cause the Lost Layer}
The results in Figure~\ref{fig:domain_influenc} of the main text demonstrate that changes in the visual domain are the primary factor causing the lost layer. To further confirm that category information is not the primary factor for the shortcut layer phenomenon, we sampled 100 categories from ImageNet each time to create 500 sub-tasks, each containing different category information. We recorded the classification performance of using the full text encoder (Full Result) and best performance achieved after removing a specific layer from the text encoder (Masked Max Result). The results, as depicted in Figure~\ref{fig:seman_inf}, indicate that the shortcut layer phenomenon is absent in nearly all sub-tasks on ImageNet.

\section{Specific Effects of the VtT Model}
The results in Figure~\ref{fig:intro_lost_ours} demonstrate that the lost layer disappears when our method is applied, indicating that our approach effectively leverages the information from the text encoder. We further elucidate the impact of our method through additional experiments.

\subsection{Enhancing Alignment}
Firstly, we measure the cosine similarity between the visual features and the corresponding textual features extracted by the model before and after applying our method. The results, shown in Figure~\ref{fig:v_t_sim}, indicate that our method increase the similarity between the visual and text features, enhancing cross-modal alignment.

\begin{figure*}[t]
\centering
\begin{minipage}[b]{0.47\linewidth}
    \centering
    \centering
    \subfloat["a photo of a Industrial Buildings."]{
    \includegraphics[width=0.24\linewidth]{./Figs/1_1.png} 
    \includegraphics[width=0.24\linewidth]{./Figs/1_2.png} 
    \includegraphics[width=0.24\linewidth]{./Figs/1_3.png} 
    \includegraphics[width=0.24\linewidth]{./Figs/1_4.png} 
    }
\end{minipage}
\medskip
\begin{minipage}[b]{0.47\linewidth}
    \centering
    \centering
    \subfloat["a photo of a Vascular Lesion."]{
    \includegraphics[width=0.24\linewidth]{./Figs/5_1.png} 
    \includegraphics[width=0.24\linewidth]{./Figs/5_2.png} 
    \includegraphics[width=0.24\linewidth]{./Figs/5_3.png} 
    \includegraphics[width=0.24\linewidth]{./Figs/5_4.png}
    }
\end{minipage}
\\
\begin{minipage}[b]{0.47\linewidth}
    \centering
    \centering
    \subfloat["a photo of a Benign Keratosi."]{
    \includegraphics[width=0.24\linewidth]{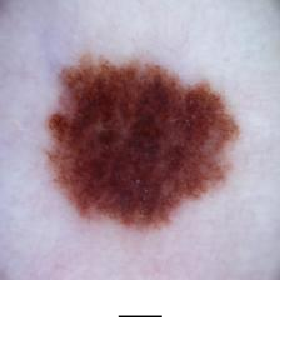} 
    \includegraphics[width=0.24\linewidth]{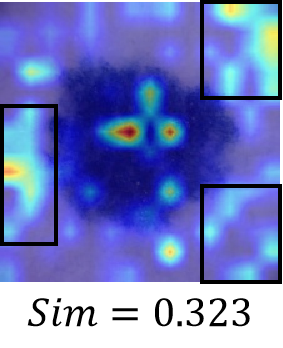} 
    \includegraphics[width=0.24\linewidth]{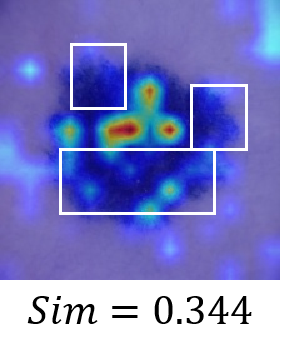} 
    \includegraphics[width=0.24\linewidth]{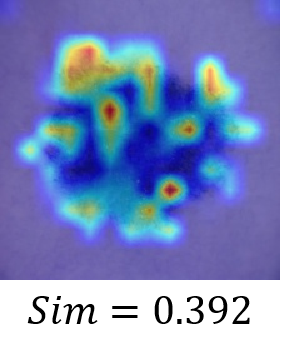}
    }
\end{minipage}
\medskip
\begin{minipage}[b]{0.47\linewidth}
    \centering
    \centering
    \subfloat["a photo of a Tomato Yellow Leaf Curl Virus."]{
    \includegraphics[width=0.24\linewidth]{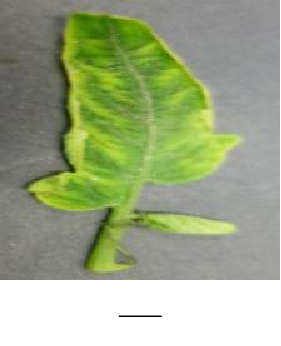} 
    \includegraphics[width=0.24\linewidth]{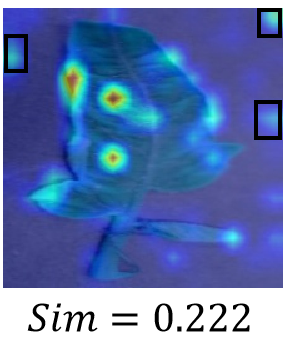} 
    \includegraphics[width=0.24\linewidth]{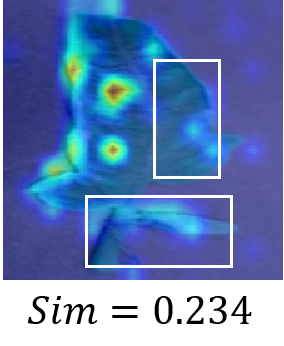} 
    \includegraphics[width=0.24\linewidth]{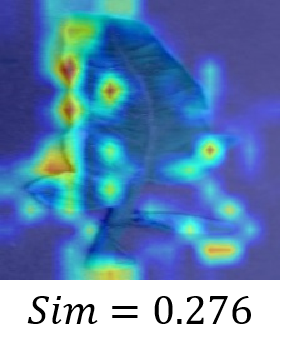}
    }
\end{minipage}
\\
\caption{The attention maps of the three models. From left to right: the original image, the baseline result, baseline + remove (see Figure 2(c) in the main text) result, and the result of ours. Black boxes highlight areas of incorrect attention, while white boxes highlight areas of missing attention. $Sim$ represents the cosine similarity between the image features and the text features. A higher similarity indicates better alignment.}
\label{fig:heat_map}
\end{figure*}

\subsection{More Domain-Independent Features}
We also visualized the features extracted by the model for images from five categories within the ImageNet-R dataset, across different domains, before and after applying our method. The results are presented in Figure~\ref{fig:tsne}. In these visualizations, colors represent the domains from which the samples originate, and different shapes denote different categories. Before applying our method, it is evident that the categories were difficult to distinguish (points of different shapes are mixed together), and samples from different domains were clearly separated (points of the same shape but different colors are distinctly divided). This indicates that the features extracted by the model were domain-dependent. However, after applying our method, we observed an improvement in classification results (points of different shapes are separated), and the model extracted consistent features for samples from the same category but different domains (points of the same shape but different colors are nearly indistinguishable). This suggests that our method enables the model to extract more domain-independent features.

\subsection{Reclaiming the Lost Layer Information}
We illustrate the specific changes in the model before and after leveraging the information from the lost layer through several examples. Figure~\ref{fig:heat_map} shows, from left to right: (1) the original image; (2) when using the full text encoder, the model incorrectly focuses on non-semantic parts (highlighted by the black box); (3) after removing the lost layer, these incorrect focuses disappear, but some effective attention areas (highlighted by the white box) are lost; (4) our method eliminates the incorrect focuses while preserving the effective attention areas, achieving an appropriate focus range and better feature alignment.

\section{Extended Results on CDFSL Task}
Table~\ref{tab:cdfsl} is an extended version of Table~\ref{tab:exp_cdfsl} in the main text. It includes the performance of various models on four CDFSL datasets under different settings. These settings involve different backbones (CLIP~\cite{radford2021learning}, SigLIP2~\cite{tschannen2025siglip}, and PE-Core~\cite{bolya2025perception}), the use of a source dataset (Source), and whether fine-tuning on the target domain is applied (FT).
Specifically, the methods include ATA~\cite{wang2021cross}, AFA~\cite{hu2022adversarial}, wave-SAN~\cite{fu2022wave}, StyleAdv~\cite{fu2023styleadv}, StyleAdv-FT (fine-tuned StyleAdv), DARA~\cite{zhao2023dual}, PMF~\cite{hu2022pushing}, FLoR~\cite{zou2024flatten}, CD-CLS~\cite{zou2025closer}, AttnTemp~\cite{zouattention}, VDB~\cite{yazdanpanah2022visual}, IM-DCL~\cite{xu2024enhancing}, StepSTP~\cite{xu2024step}, CoOp~\cite{zhou2022learning}, Tip-Adapter~\cite{zhang2021tip}, AMU-Tuning~\cite{tang2024amu}, LP++~\cite{huang2024lp++}, LDC~\cite{li2025logits}, Maple~\cite{khattak2023maple}, and CLIP-LoRA~\cite{zanella2024low}, which are introduced as our competitors.

\section{Few Shot Learning Results on Meta-dataset}
We further demonstrate the effectiveness of our method on the Meta-dataset~\cite{triantafillou2019meta}. We compare our method with the most recent CLIP-based methods and present the results in Table~\ref{tab:meta_result}. For results on datasets ISIC, EuroSAT, ChestX, and CropDisease, please refer to Table~\ref{tab:cdfsl}. This comparison includes both fully fine-tuned methods~\cite{song2023fd,wortsman2022robust} and prompt-tuning methods~\cite{khattak2023maple,brahma2025prompt} of CLIP. As shown in Table~\ref{tab:meta_result}, our method, when added to CLIP-LoRA~\cite{zanella2024low}, effectively enhances its performance, achieving an improvement of approximately 3 points in the 1-shot scenario. Additionally, CLIP-LoRA + OURS achieves the highest average performance to date in both 1-shot and 5-shot tasks.
\begin{table*}[htbp]
\belowrulesep=0pt
\aboverulesep=0pt
\caption{
The accuracy(\%) of four target domain datasets under 5-way 1-shot and 5-way 5-shot tasks. The use of a source dataset (Source), and whether fine-tuning on the target domain is applied (FT)}
\label{tab:cdfsl}    
\centering
\begin{adjustbox}{max width=0.90\linewidth}
\begin{tabular}{*{10}{c|c c|c c c c c c c|} }  
\toprule 
 \multirow{1}*{Task}  &  \multirow{1}*{Method}  &\multirow{1}*{backbone} &  \multirow{1}*{Source} &  \multirow{1}*{FT}  & \multirow{1}*{ISIC} & \multirow{1}*{EuroSAT} & \multirow{1}*{CropDisease} & \multirow{1}*{ChestX} & \multirow{1}*{Avg} \\           
 \midrule
  &ATA~\cite{wang2021cross}  &RN10 &Y &-  &33.21±0.40 &61.35±0.50 &67.47±0.50 &22.10±0.20 &46.03 \\
  &AFA~\cite{hu2022adversarial}  &RN10 &Y &-  &33.21±0.30 &63.12±0.50 &67.61±0.50 &22.92±0.20 &46.72 \\
  &wave-SAN~\cite{fu2022wave}  &RN10 &Y &- &33.35±0.71 &69.64±1.09 &70.80±1.06 &22.93±0.49 &49.18 \\
  &StyleAdv~\cite{fu2023styleadv}  &RN10 &Y &- &33.96±0.57 &70.94±0.82 &74.13±0.78 &22.64±0.35 &50.42 \\
  &ATA-FT~\cite{wang2021cross}  &RN10 &Y &Y &34.94±0.40 &68.62±0.50 &75.41±0.50 &22.15±0.20 &50.28 \\
  &DARA~\cite{zhao2023dual}  &RN10 &Y &Y &36.42±0.64 &67.42±0.80 &80.74±0.76 &22.92±0.40 &51.88 \\
  &StyleAdv-FT~\cite{fu2023styleadv}  &RN10 &Y &Y &35.76±0.52 &72.92±0.75 &80.69±0.28 &22.64±0.35 &53.00 \\
  \cline{2-10}
  \multirow{14}{*}{\rotatebox{90}{5-way 1-shot}} 
  &PMF~\cite{hu2022pushing}  &ViT/DINO &Y &Y &30.36±0.36 &70.74±0.63 &80.79±0.62 &21.73±0.30 &50.91 \\
  &StyleAdv-FT~\cite{fu2023styleadv} &ViT/DINO &Y &Y &33.99±0.46 &74.93±0.58 &84.11±0.57 &22.92±0.32 &53.99 \\
  &FLoR~\cite{zou2024flatten} &ViT/DINO &Y &Y &35.49 &73.09 &83.55 &23.26 & 53.85 \\
  &CD-CLS~\cite{zou2025closer} &ViT/DINO &Y &Y  &35.56 & 74.97 & 84.53 &23.39 & 54.62 \\
  &AttnTemp~\cite{zouattention} &ViT/DINO &Y &Y  &38.05 &75.09 & 84.78 &23.63 & 55.39 \\
  \cline{2-10}
  &FN+VDB~\cite{yazdanpanah2022visual}  &RN18 &- &Y &32.96±0.57 &69.67±0.80 &79.68±0.74 &22.64±0.41 &51.24 \\
  &IM-DCL~\cite{xu2024enhancing}  &RN10 &- &Y &38.13±0.57 &77.14±0.71 &84.37±0.99 &23.98±0.79 &55.91 \\
  \cline{2-10}
  &StepSTP~\cite{xu2024step} &ViT/CLIP &- &Y &32.97±0.27 &70.01±0.21 &84.84±0.72 &\textbf{22.84±0.95} &52.68 \\
  &CoOp~\cite{zhou2022learning}  &ViT/CLIP &- &Y  &32.86±0.47 &72.08±0.66 &80.50±0.74 &21.65±0.32 &51.77 \\
  &Tip-Adapter~\cite{zhang2021tip}  &ViT/CLIP &- &Y  &32.68±0.37 &75.44±0.51 &77.15±0.66 &22.24±0.26 &51.87 \\
  &PromptSRC~\cite{Khattak_2023_ICCV}  &ViT/CLIP &- &Y  &31.86±0.57 &73.44±0.71 &76.15±0.89 &21.16±0.36 &50.65 \\
  &PDA~\cite{bai2024prompt} &ViT/CLIP &- &Y &31.45±0.44 &69.68±0.71 &79.20±0.83 &20.66±0.28 &50.25\\
  &AMU-Tuning~\cite{tang2024amu}  &ViT/CLIP &- &Y  &32.29±0.67 &72.24±0.71 &80.20±0.86 &21.56±0.36 &51.57 \\
  &LP++~\cite{huang2024lp++}  &ViT/CLIP &- &Y  &33.63±0.41 &73.05±0.55 &81.84±0.66 &21.72±0.42 &52.56 \\
  &LDC~\cite{li2025logits}  &ViT/CLIP &- &Y  &33.72±0.46 &74.19±0.52 &83.77±0.81 &22.12±0.36 &53.45 \\
  &Maple~\cite{khattak2023maple}  &ViT/CLIP &- &Y &33.38±0.49 &76.05±0.63 &81.78±0.72 &21.09±0.31 &53.07 \\
  \rowcolor{cyan!10} \cellcolor{white}
  &\textbf{Maple + OURS}  &ViT/CLIP &- &Y &34.06±0.53 &82.25±0.75 &82.66±0.72 &21.64±0.34 &55.15 \\
  &CLIP-LoRA-Vision~\cite{zanella2024low}  &ViT/CLIP &- &Y &36.40±0.42 &81.72±0.52 &84.22±0.62 &21.86±0.32 &55.97 \\
  \rowcolor{cyan!10} \cellcolor{white}
  &\textbf{CLIP-LoRA-Vision + OURS}  &ViT/CLIP &- &Y &\textbf{38.20±0.45} &\textbf{85.01±0.41} &\textbf{87.00±0.53} &22.70±0.33 &\textbf{58.23} \\
  \cline{2-10}
  &SigLIP2-LoRA~\cite{tschannen2025siglip} &ViT/SigLip2 &- &Y  &33.47 &74.16 & 87.50 &21.44 & 54.14 \\
  \rowcolor{cyan!10} \cellcolor{white}
  &\textbf{SigLIP2-LoRA + OURS} &ViT/SigLip2 &- &Y  &\textbf{35.34} &\textbf{76.10} & \textbf{89.72} &\textbf{22.00} & \textbf{55.79} \\
  \cline{2-10}
  &PE-Core-LoRA~\cite{bolya2025perception} &ViT/PE-Core &- &Y  &40.89 &84.49 & 91.75 &22.02 & 59.78 \\
  \rowcolor{cyan!10} \cellcolor{white}
  &\textbf{PE-Core-LoRA + OURS} &ViT/PE-Core &- &Y  &\textbf{42.20} &\textbf{86.16} & \textbf{92.61} &\textbf{23.04} & \textbf{61.00} \\
  \midrule
  \midrule
  &ATA~\cite{wang2021cross} &RN10 &Y &- &44.91±0.40 &83.75±0.40 &90.59±0.30 &24.32±0.40 & 60.89 \\
  &AFA~\cite{hu2022adversarial} &RN10 &Y &-  &46.01±0.40 &85.58±0.40 &88.06±0.30 &25.02±0.20 &61.17 \\
  &wave-SAN~\cite{fu2022wave} &RN10 &Y &-  &44.93±0.67 &85.22±0.71 &89.70±0.64 &25.63±0.49 &61.37 \\
  &StyleAdv~\cite{fu2023styleadv} &RN10 &Y &- &45.77±0.51 &86.58±0.54 &93.65±0.39 &26.07±0.37 &63.02 \\
  &ATA-FT~\cite{wang2021cross} &RN10 &Y &Y &49.79±0.40 &89.64±0.30 &95.44±0.20 &25.08±0.20 &64.99 \\
  &DARA~\cite{zhao2023dual} &RN10 &Y &Y &56.28±0.66 &85.84±0.54 &95.32±0.34 &27.54±0.42 &66.25 \\
  &StyleAdv-FT~\cite{fu2023styleadv} &RN10 &Y &Y &53.05±0.54 &91.64±0.43 &96.51±0.28 &26.24±0.35 &66.86 \\
  \cline{2-10}
  \multirow{14}{*}{\rotatebox{90}{5-way 5-shot}} 
  &PMF~\cite{hu2022pushing} &ViT/DINO &Y &Y &50.12 &85.98 &92.96 &27.27 &64.08 \\
  &StyleAdv-FT~\cite{fu2023styleadv} &ViT/DINO &Y &Y &51.23±0.51 &90.12±0.33 &95.99±0.27 &26.97±0.33 &66.08 \\
  &FLoR~\cite{zou2024flatten} &ViT/DINO &Y &Y &53.06 &90.75 &96.47 & 27.02 & 66.83 \\
  &CD-CLS~\cite{zou2025closer} &ViT/DINO &Y &Y  &54.69 & 91.53 & 96.27 &27.66 & 67.54 \\
  &AttnTemp~\cite{zouattention} &ViT/DINO &Y &Y  &54.91 &90.82 & 96.66 &28.03 & 67.61 \\  
  \cline{2-10}
  &FN+VDB~\cite{yazdanpanah2022visual} &RN18 &- &Y &47.48±0.59 &87.31±0.50 &94.63±0.37 &25.55±0.43 &64.74 \\
  &IM-DCL~\cite{xu2024enhancing} &RN10 &- &Y &52.74±0.69 &89.47±0.42 &95.73±0.38 &28.93±0.41 &66.72 \\
  \cline{2-10}
  &StepSTP~\cite{xu2024step} &ViT/CLIP &- &Y &52.12±0.36 &89.40±1.05 &96.01±0.88 &26.36±0.97 &65.97 \\
  &CoOp~\cite{zhou2022learning}  &ViT/CLIP &- &Y &45.78±0.75 &85.88±0.49 &93.31±0.57 &23.35±0.50 &62.08 \\
  &Tip-Adapter~\cite{zhang2021tip}  &ViT/CLIP &- &Y  &46.96±0.59 &87.24±0.33 &94.19±0.39 &24.07±0.44 &63.12 \\
  &PromptSRC~\cite{Khattak_2023_ICCV}  &ViT/CLIP &- &Y  &46.09±0.48 &86.54±0.49 &89.97±0.41 &23.51±0.47 &61.52 \\
  & PDA~\cite{bai2024prompt} &ViT/CLIP &- &Y &45.19±0.62 &86.21±0.44 &92.67±0.39 &21.87±0.33 &61.48 \\
  &AMU-Tuning~\cite{tang2024amu}  &ViT/CLIP &- &Y  &44.60±0.62 &88.47±0.39 &94.26±0.52 &23.34±0.41 &62.66 \\
  &LP++~\cite{huang2024lp++}  &ViT/CLIP &- &Y  &48.49±0.44 &87.48±0.42 &94.47±0.38 &23.89±0.29 &63.58 \\
  &LDC~\cite{li2025logits}  &ViT/CLIP &- &Y  &49.70±0.33 &90.82±0.22 &96.71±0.34 &25.89±0.21 &65.78 \\
  &Maple~\cite{khattak2023maple}  &ViT/CLIP &- &Y &48.35±0.75 &89.04±0.52 &93.50±0.54 &22.96±0.50 &63.46 \\
  \rowcolor{cyan!10} \cellcolor{white}
  &\textbf{Maple + OURS}  &ViT/CLIP &- &Y &49.81±0.78 &92.24±0.42 &94.62±0.53 &24.04±0.50 &65.18 \\
  &CLIP-LoRA-Vision~\cite{zanella2024low}  &ViT/CLIP &- &Y &52.22±0.71 &93.31±0.47 &95.88±0.42 &24.61±0.47 &66.50 \\
  \rowcolor{cyan!10} \cellcolor{white}
  &\textbf{CLIP-LoRA-Vision + OURS}  &ViT/CLIP &- &Y &\textbf{56.20±0.41} &\textbf{94.58±0.31} &\textbf{97.21±0.35} &\textbf{26.42±0.31} &\textbf{68.57}\\
  \cline{2-10}
  &SigLIP2-LoRA~\cite{tschannen2025siglip} &ViT/SigLip2 &- &Y  &51.79 &91.39 & 96.43
  &24.24 & 65.96 \\
  \rowcolor{cyan!10} \cellcolor{white}
  &\textbf{SigLIP2-LoRA + OURS} &ViT/SigLip2 &- &Y  &\textbf{55.11} &\textbf{92.70} & \textbf{97.63} &\textbf{25.54} & \textbf{67.75} \\
  \cline{2-10}
  &PE-Core-LoRA~\cite{bolya2025perception} &ViT/PE-Core &- &Y  &58.81 &94.07 & 97.25 &24.44 & 68.64 \\
  \rowcolor{cyan!10} \cellcolor{white}
  &\textbf{PE-Core-LoRA + OURS} &ViT/PE-Core &- &Y  &\textbf{60.03} &\textbf{94.67} & \textbf{98.36} &\textbf{27.05} & \textbf{70.05} \\
\bottomrule
\end{tabular}
\end{adjustbox}
\end{table*}

\begin{table*}[!htbp]
\caption{
Detailed results for meta-dataset~\cite{triantafillou2019meta} with the ViT-B/16 as visual backbone. Highest value is high lighted inbold.
}
\label{tab:meta_result}    
\centering
\begin{adjustbox}{max width=0.98\linewidth}
\begin{tabular}{*{15}{c}}  
\toprule 
 \multirow{1}*{Shots} & \multirow{1}*{Method}  &  \multirow{1}*{Omniglot}  & \multirow{1}*{Traffic Signs}
 & \multirow{1}*{MSCOCO}  & \multirow{1}*{Textures}  & \multirow{1}*{CUB}  
  & \multirow{1}*{Quickdraw}   & \multirow{1}*{Aircraft}  & \multirow{1}*{VGG Flower}  
  & \multirow{1}*{Fungi}  & \multirow{1}*{Mini-test} & \multirow{1}*{Average}\\
\midrule
\multirow{6}{*}{\rotatebox{90}{1-shot}}
&WiSE-FT &83.56&60.84&67.28&63.55&81.16&62.54&62.64&73.14&59.10&93.55&70.73\\
&FD-Align &83.81&57.32&65.91&66.05&82.88&64.49&62.90&79.87&57.05&95.04&71.53\\ 
&MaPLe &77.82&56.45&68.55&66.05&94.08&72.54&79.76&98.24&55.85&97.17&76.65\\
&PromptMargin &87.01&67.24&72.86&69.25&\textbf{96.97}&74.84&83.94&\textbf{98.43}&61.16&\textbf{99.17}&81.08\\
&CLIP-LoRA &90.29&78.45&83.45&84.75&93.90&76.08&80.83&97.57&64.59&98.97&84.88\\
&CLIP-LoRA + OURS &\textbf{95.01}&\textbf{85.40}&\textbf{83.89}&\textbf{86.27}&\underline{95.97}&\textbf{82.37}&\textbf{84.02}&97.94&\textbf{66.68}&\underline{98.99}&\textbf{87.64}\\

\bottomrule
\multirow{6}{*}{\rotatebox{90}{5-shot}}
&WiSE-FT &95.26&78.11&81.08&83.31&93.41&82.78&77.66&99.06&73.28&98.44&86.24\\
&FD-Align &94.81&73.39&81.37&83.60&93.87&82.78&78.21&98.95&73.69&98.52&85.91\\
&Maple &96.23&85.21&75.13&88.45&97.65&85.08&87.56&99.23&79.69&\textbf{99.39}&89.06\\
&PromptMargin &96.37&87.55&76.68&88.71&97.12&85.21&86.53&99.27&80.91&99.19&89.75\\
&CLIP-LoRA &98.78&94.50&86.61&90.66&96.60&89.33&87.44&99.29&82.68&99.10&92.49\\
&CLIP-LoRA + OURS &\textbf{99.40}&\textbf{96.73}&\textbf{86.67}&\textbf{91.06}&\textbf{97.83}&\textbf{90.11}&\textbf{87.63}&\textbf{99.51}&\textbf{84.18}&\underline{99.20}&\textbf{93.22}\\

\bottomrule
\end{tabular}
\end{adjustbox}
\end{table*}

\section{Detailed Implementation Details.}
We adopt the ViT-Base/16 network as the backbone, utilizing parameters pre-trained by CLIP~\cite{radford2021learning}, SigLIP2~\cite{tschannen2025siglip} and PE-Core~\cite{bolya2025perception}. In every layer of the visual branch, we incorporate the LoRA~\cite{hu2021lora} structure for fine-tuning. For each layer's LoRA component, we set $r = 16$ and $\alpha = 8$. Following the methodology in \cite{xu2024step}, we perform data augmentation on the support samples in each episode. Subsequently, we train the model for 250 epochs, using two hyperparameters, $\beta$ and $\lambda$ (refer to Method). And for all settings, we fix $\beta = 7$ and $\lambda = 50$. We evaluate each model using 15 query samples per class, randomly selecting 800 episodes, and report the results (in percentage) with a 95\% confidence interval. All training and testing procedures are conducted on a single NVIDIA GeForce RTX 3090.

\section{Detailed Related Work}
\subsection{Source-Free Cross-domain Few shot Learning}
 Cross-Domain Few-Shot Learning (CDFSL) aims to train a model on a source domain that can generalize effectively to a target domain with limited examples. Existing methods are typically categorized into two types: meta-learning-based approaches~\cite{fu2022wave,guo2020broader,hu2022adversarial,wang2021cross} and transfer learning-based approaches~\cite{guo2020broader,liang2021boosting,zhou2023revisiting,zou2024flatten,zou2025closer,zouattention}. Source-Free Cross-Domain Few-Shot Learning (SF-CDFSL) introduces a stronger constraint by making source domain data inaccessible. Current SF-CDFSL methods~\cite{yazdanpanah2022visual,zhuo2024prompt,xu2024step} primarily rely on large models, such as CLIP~\cite{radford2021learning}, leveraging their prior knowledge for classification in the target domain. However, no prior work has identified or analyzed the issue of the lost layer when using CLIP for CDFSL tasks.
\subsection{Parameter-Efficient Fine-Tuning}
A crucial area of research is the efficient application of Vision-Language Models (VLMs) to downstream tasks. A widely used approach to achieve this is parameter-efficient fine-tuning (PEFT), which involves utilizing only a small number of samples from the target task. PEFT focuses on adjusting a limited subset of the VLM’s parameters, enabling the model to adapt to diverse applications without altering all pre-trained parameters. There are three main categories of PEFT methods: prompt learning, adapters, and LoRA (along with its variants). Prompt learning involves converting fixed templates into learnable parameters, as demonstrated in works like CoOp~\cite{zhou2022learning}, CoCoOp~\cite{zhou2022conditional}, MaPLe~\cite{khattak2023maple}, PLOT~\cite{chen2022plot}, ProGrad~\cite{zhu2023prompt}, PromptSRC~\cite{khattak2023self}, KgCoOp~\cite{yao2023visual}, PCB~\cite{bang2024active}, DynaPrompt~\cite{zehao2025dynaprompt}, TCP~\cite{yao2024tcp}, and ATPrompt~\cite{li2025advancing}. Moreover, Customized Ensemble~\cite{lu2023beyond} improves performance by combining outputs from multiple models, while PromptKD~\cite{li2024promptkd} explores knowledge distillation within prompt learning. Adapter-based methods, such as CLIP-Adapter~\cite{gao2024clip}, Tip-Adapter~\cite{zhang2022tip}, LP++~\cite{huang2024lp++}, AMU-Tuning~\cite{tang2024amu}, LatHAdapter~\cite{zhao2025fine}, MMA~\cite{Yang_2024_CVPR} and LDC~\cite{li2025logits}. Low-Rank Adaptation (LoRA)~\cite{hu2021lora,zanella2024low} fine-tunes the model by adding learnable low-rank matrices while keeping the original parameters fixed. The new weights can be merged with the original ones, and LoRA does not add extra inference time. Various studies have extended LoRA by adapting the rank for each matrix~\cite{valipour2022dylora,zhang2023adalora}, improving its performance~\cite{chavan2023one,kim2024hydra,zi2023delta}, or reducing memory usage through quantization~\cite{dettmers2024qlora,rajabzadeh2024qdylora}

\subsection{Layer Redundancy}
Research related to layer redundancy, such as \cite{lad2406remarkable,gonzalez2025leveraging} in LLMs, differs from ours. We focus on VLMs, such as CLIP~\cite{radford2021learning}, and examine this phenomenon in the SF-CDFSL task setting. Unlike previous studies \cite{lawson2025learning,tong2025flowcut,wang2025investigating,men2024shortgpt} which focus on in-domain scenarios, consider these layers as redundant, and employ a removal strategy, we study the SF-CDFSL problem on VLM and find the information in these layers actually beneficial for SF-CDFSL tasks, instead of removing layers~\cite{men2024shortgpt,tong2025flowcut,lad2406remarkable}, we reclaim the lost layer and demonstrate that this reclamation is a superior strategy compared to removal. Our work provides a new perspective for analyzing similar issue. 

\section{Broader Impact}
In this paper, we observe the Lost Layer in CLIP under cross-domain scenarios and reveal that the information within the Lost Layer is actually beneficial for SF-CDFSL tasks. However, changes in the visual domain lead to the under-utilization of this information. We then propose the VtT model to reutilize this information. Extensive experiments validate our rationale and effectiveness. Our research is crucial for future studies on fine-tuning VLM models in cross-domain scenarios. From the perspective of shortcut learning, it highlights the impact of the Lost Layer and discusses how to effectively utilize it. While our method has been evaluated across four distinct target domains, offering a promising initial assessment of its cross-domain applicability, the diversity of these domains may not fully capture all potential real-world scenarios. Future work will focus on expanding our evaluations to include a broader range of target domains to better understand the method's performance in diverse real-world contexts.